\documentclass[runningheads]{llncs}

\usepackage{eccv}

\usepackage{eccvabbrv}

\usepackage{graphicx}
\usepackage{booktabs}
\usepackage{enumitem}
\usepackage{tcolorbox}
\usepackage{xcolor}
\usepackage{multirow}
\usepackage{adjustbox}
\usepackage{pifont}
\usepackage{placeins}

\usepackage[accsupp]{axessibility}  

\usepackage{hyperref}

\usepackage{orcidlink}

\newcommand{\commandnotation}[1]{\textless \textit{#1}\textgreater}
\definecolor{mydarkgreen}{RGB}{30,130,30}

\begin{document}

\title{Pointer-CAD v2: Plan-Then-Construct CAD Generation with Dimension-Aware Parametric Precision}

\titlerunning{Pointer-CAD v2}

\author{
  Dacheng Qi\inst{1,2,3}\orcidlink{0009-0009-9919-8662} \and
  Chenyu Wang\inst{1,2,3} \and
  Jingwei Xu\inst{4}\orcidlink{0009-0009-2637-6632} \and
  Yi Ma\inst{1,2,3,5}\orcidlink{0000-0001-5485-419X} \and
  Shenghua Gao\raisebox{0.2ex}{\thanks{Corresponding author: \email{gaosh@hku.hk}}}\inst{,1,2,3}\orcidlink{0000-0003-1626-2040}
}

\authorrunning{D.~Qi et al.}

\institute{
  The University of Hong Kong, Hong Kong SAR \and
  Shenzhen Loop Area Institute, Shenzhen, China \and
  TranscEngram, Shenzhen, China \and
  Monash University, Melbourne, Australia \and
  University of California, Berkeley, USA
}

\maketitle

\begin{abstract}
Computer-aided design (CAD) plays a fundamental role in modern manufacturing by providing the high precision required for industrial production.
Recent large language model based approaches formulate CAD generation as a sequence prediction problem and have achieved promising results.
However, existing methods and evaluation protocols primarily emphasize visual similarity, while overlooking precise geometric parameters and correct metric scale.
Small numerical deviations that are negligible at the shape-level may still violate industrial tolerance requirements, a problem further compounded by current autoregressive paradigms that utilize command sequence representations, aggressively quantize numerical parameters to ease LLM prediction.
In this work, we present \textbf{Pointer-CAD v2}. Compared with \textbf{v1} \cite{qi2026pointer}, this version directly predicts continuous values, bypassing the need for quantized numerical parameters and thereby eliminating quantization errors. Specifically, we propose a unified framework that decouples parameter reasoning from geometric construction through a \textbf{Plan-Then-Construct paradigm}.
Our method first produces a structured design plan with explicit metric scale parameters.
These parameters are organized into a dictionary and directly referenced during sequence generation via a pointer mechanism, eliminating discretization errors and ensuring dimensionally consistent execution.
In addition, we construct a new large-scale dataset with plan-level annotation and introduce three hierarchical geometry accuracy metrics to evaluate parametric fidelity at the vertex, edge, and face levels.
Extensive experiments demonstrate that Pointer-CAD v2 consistently outperforms existing baselines and achieves substantial improvements in geometric accuracy, enabling reliable CAD generation for precision-critical engineering applications.
Our code is available at \url{https://github.com/Snitro/Pointer-CAD-v2}.


\end{abstract}

\section{Introduction}

\begin{figure}[tb]
  \centering
  \includegraphics[width=\linewidth]{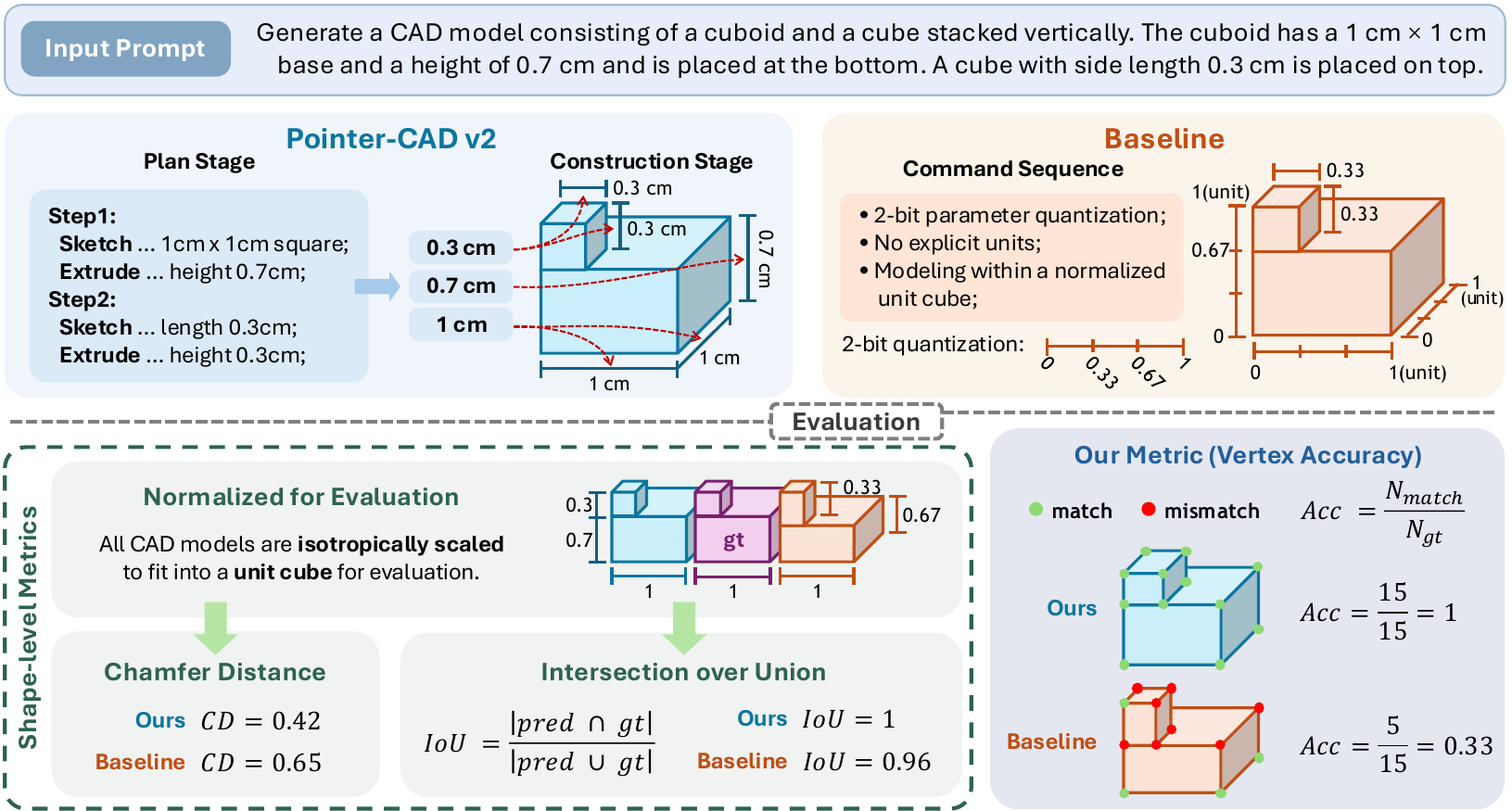}
  \caption{ 
    \textbf{Strength of Our Proposed Method and Metric.}
    Pointer-CAD v2 generates CAD models with full-precision parameters and dimensional consistency by explicitly separating parameter reasoning from command construction.
    In contrast, command-sequence baselines rely on quantization and cannot accurately represent precise dimensions.
    Unlike shape-level metrics such as Chamfer Distance and Intersection over Union, our proposed metric spatially measures parametric correctness and is more sensitive to dimensional errors rather than visual similarity.
  }
  \label{fig:teaser}
\end{figure}

Computer-Aided Design (CAD) plays a fundamental role in modern industrial design and manufacturing, enabling simulation and fabrication workflows under strict, near-zero tolerance precision requirements.
Recent advances in large-scale CAD datasets \cite{wu2021deepcad, chen2025img2cad, dong2025histcad} and generative modeling \cite{karras2022elucidating, du2023reduce, esser2024scaling} have enabled automatic synthesis of parametric CAD programs \cite{le2025cadknitter, li2025caddreamer, elistratov2026cadevolve}, yet ensuring accurate and dimensionally consistent parameters remains an open challenge, as errors in dimensional values can severely impact production-level engineering processes.

Existing methods \cite{jones2025solver, schupbach2025text, zhou2025cadialogue, qin2025drawing2cad} formulate CAD generation as sequence prediction under an autoregressive paradigm and can be broadly categorized into two groups: command sequence representations and code representations.
Code representations generate programs in scripting environments such as CADQuery \cite{xie2025text} or the FreeCAD API \cite{mallis2025cad, niu2025creft}.
They align naturally with large language models (LLMs) \cite{yang2025qwen3, team2026kimi} and programming ecosystems \cite{cadquery, freecad}, but generating a single model is often token-intensive due to the verbosity of uncompressed code \cite{qi2026pointer}. 
In contrast, command sequence representations \cite{xu2024cad} use specialized token vocabularies to describe parametric operations, offering higher token efficiency.
However, they remain largely limited to sketch-extrude pair operations and struggle with high-precision modeling due to the quantization of dimensions introduced to simplify autoregressive generation.
Recently, Pointer-CAD \cite{qi2026pointer} extends the representation capability of command sequence models through explicit entity referencing, enabling support for more diverse CAD operations.
Precise dimensional control, however, remains unresolved and limits its applicability in industrial and precision-sensitive scenarios.

The design of command sequence representations inherently constrains precise geometric and dimensional accuracy, since parameters and geometric operations share the same tokenized space \cite{wu2021deepcad, qi2026pointer}.
As a result, continuous geometric quantities must be quantized into a finite vocabulary, leading to a loss of parametric fidelity.
Existing evaluation metrics further obscure this issue. 
Widely adopted metrics such as Chamfer Distance (CD) and Intersection over Union (IoU), illustrated in \cref{fig:teaser}, mainly measure shape-level similarity and are insensitive to small numerical deviations. 
However, such deviations are unacceptable in industrial CAD settings that require strict dimensional tolerances.
Although Pointer-CAD~\cite{qi2026pointer} mitigates quantization errors through its pointer mechanism, particularly for sketch plan localization and special edge endpoint positioning (e.g., midpoints or existing edge endpoints), small numerical deviations may still arise in other unconstrained geometric configurations.
These deviations are subtle and are not explicitly captured by shape-level evaluation metrics.
As a result, neither existing methods nor evaluation metrics are explicitly optimized for accurate metric scale, which limits their applicability in precision-critical engineering design.
To address these limitations, we propose a unified framework that decouples parameter reasoning from geometric construction while preserving accurate parameters together with their associated units.
Our method, termed \textbf{Pointer-CAD v2}, extends Pointer-CAD with a structured Plan-Then-Construct paradigm.
Pointer-CAD generates CAD models in a step-wise manner and employs a pointer mechanism to reference intermediate geometric entities.
Our method augments each step with a plan stage that produces a structured, dimension-aware design plan.
The parameters defined in the plan are extracted to form a parameter dictionary, which is then used to generate the command sequence.
Instead of predicting parameters in a quantized space, a pointer-based mechanism directly selects values from the parameter dictionary during sequence generation.
This design eliminates discretization errors, decouples numerical reasoning from geometric construction, and ensures dimensionally consistent parameter usage.

To enable structured parameter planning with explicit unit grounding and rigorous evaluation, we construct a new dataset and introduce geometry accuracy metrics.
Built upon Recap-OmniCAD and Recap-OmniCAD\textsuperscript{+}~\cite{qi2026pointer}, where the latter additionally includes models with \textit{chamfer} and \textit{fillet}, we establish plan-level supervision by annotating structured design plans for each CAD model using Qwen3~\cite{yang2025qwen3}.
This results in OmniCAD-Plan with 202K unique valid models and OmniCAD-Plan\textsuperscript{+} with unique 209K valid models. 
We further propose geometry accuracy metrics that spatially measure parametric correctness rather than visual similarity. 
The evaluation protocol verifies whether generated models satisfy the intended dimensional constraints within predefined tolerances at three geometric levels, including vertex, edge, and face. 
As illustrated in \cref{fig:teaser}, these metrics are more sensitive to dimensional deviations and better distinguish visually similar models with different numerical parameters.
Under the proposed dataset and evaluation protocol, Pointer-CAD v2 significantly outperforms existing baselines across three geometric levels, demonstrating superior parametric accuracy and dimensional consistency.

Our contributions are summarized as follows: (1) We propose a Plan-Then-Construct framework for CAD generation and introduce Pointer-CAD v2, which explicitly separates parameter reasoning from geometric construction and enables the generation of dimensionally accurate CAD models; (2) We introduce a new dataset and hierarchical metrics for training and evaluating parametric fidelity beyond visual similarity; (3)Extensive experiments demonstrate that Pointer-CAD v2 outperforms baseline methods by eliminating the need to predict quantized numerical parameters, improving the dimensional precision.

\section{Related Work}

\subsection{Command Sequence Representations}

Command sequence representations have become a popular paradigm for CAD generation, largely enabled by the availability of large-scale and diverse datasets~\cite{wu2021deepcad, willis2021fusion, khan2024text2cad}.
Early work such as DeepCAD~\cite{wu2021deepcad} formulates CAD modeling as sequence prediction over parametric operations, while subsequent methods improve structural modeling capacity through hierarchical designs \cite{xu2022skexgen, xu2023hierarchical} and text-conditioned generation \cite{khan2024text2cad}.
Recent advances further enhance generation fidelity with more expressive architectures, including diffusion-based models~\cite{ma2024draw, zhang2025diffusion, yu2026gencad} and LLMs~\cite{wang2025cad, li2025seek, zheng2025target}.
In parallel, reverse-engineering approaches aim to translate alternative representations, including B-rep~\cite{yin2025rlcad, zhang2025ecad, liang2025cadcl}, point clouds~\cite{dupont2024transcad, khan2024cad, ma2024draw}, and images~\cite{wu2024cadvlm, chen2025img2cad, chen2025cadcrafter}, into executable command sequences.
CAD-MLLM~\cite{xu2024cad} further unifies multiple modalities within a single framework.
Pointer-CAD~\cite{qi2026pointer} introduces a pointer mechanism for explicit geometric referencing, expanding the expressive capacity of command sequence.
Despite these advances, most existing methods operate in normalized and quantized parameter spaces, leading to discretization errors and missing explicit metric dimension information.

\subsection{Code Representations}

With the rapid progress of LLMs in code generation~\cite{chen2021evaluating, zheng2023codegeex}, code-based representations have become an alternative paradigm for CAD modeling.
Early studies often designed domain-specific languages for CAD representation~\cite{fan2025parametric, govindarajan2026cadmium}, while recent works adopt mature CAD scripting environments such as CadQuery~\cite{guan2025cad, li2025cad, niu2025creft, xie2025text}, PythonOCC~\cite{yuan2024openecad}, and FreeCAD~\cite{mallis2025cad, niu2025creft}.
By grounding generation in executable APIs, these approaches leverage the coding capabilities of pre-trained LLMs.
Code representations have also been extended beyond text-to-CAD synthesis to reverse engineering from point clouds or images~\cite{rukhovich2025cad, you2025img2cad}, as well as multi-modal and agent-based frameworks~\cite{kolodiazhnyi2025cadrille, mallis2025cad}.
Leveraging mature CAD scripting environments, code-based methods can naturally represent high-precision parameters with explicit dimensions.
However, they typically require around four times more tokens than command sequence representations \cite{qi2026pointer}, reducing both training and inference efficiency.

\subsection{Planning-Based CAD Generation}

Planning-based CAD generation seeks to decouple high-level design reasoning from low-level geometric construction~\cite{guo2026plan}. 
Prior works introduce intermediate planning stages to improve interpretability and long-horizon reasoning. 
For instance,~\cite{li2026draw} represent 2D sketching as sequential geometric constructions, while FreeCAD~\cite{lin2025freecad} and CAD-Assistant~\cite{mallis2025cad} generate structured plans to guide subsequent tool calls. 
Seek-CAD~\cite{li2025seek} integrates visual feedback with structured reasoning for iterative self-refinement. 
In parallel, concurrent works~\cite{wang2023recad, guan2025cad, niu2025intent} incorporate Chain of Thought (CoT) reasoning and reinforcement learning to enhance long-horizon generation.
However, these approaches do not establish an explicit linkage between planning outputs and metric scale parameters.

\section{Method}

\begin{figure}[tb]
  \centering
  \includegraphics[width=\linewidth]{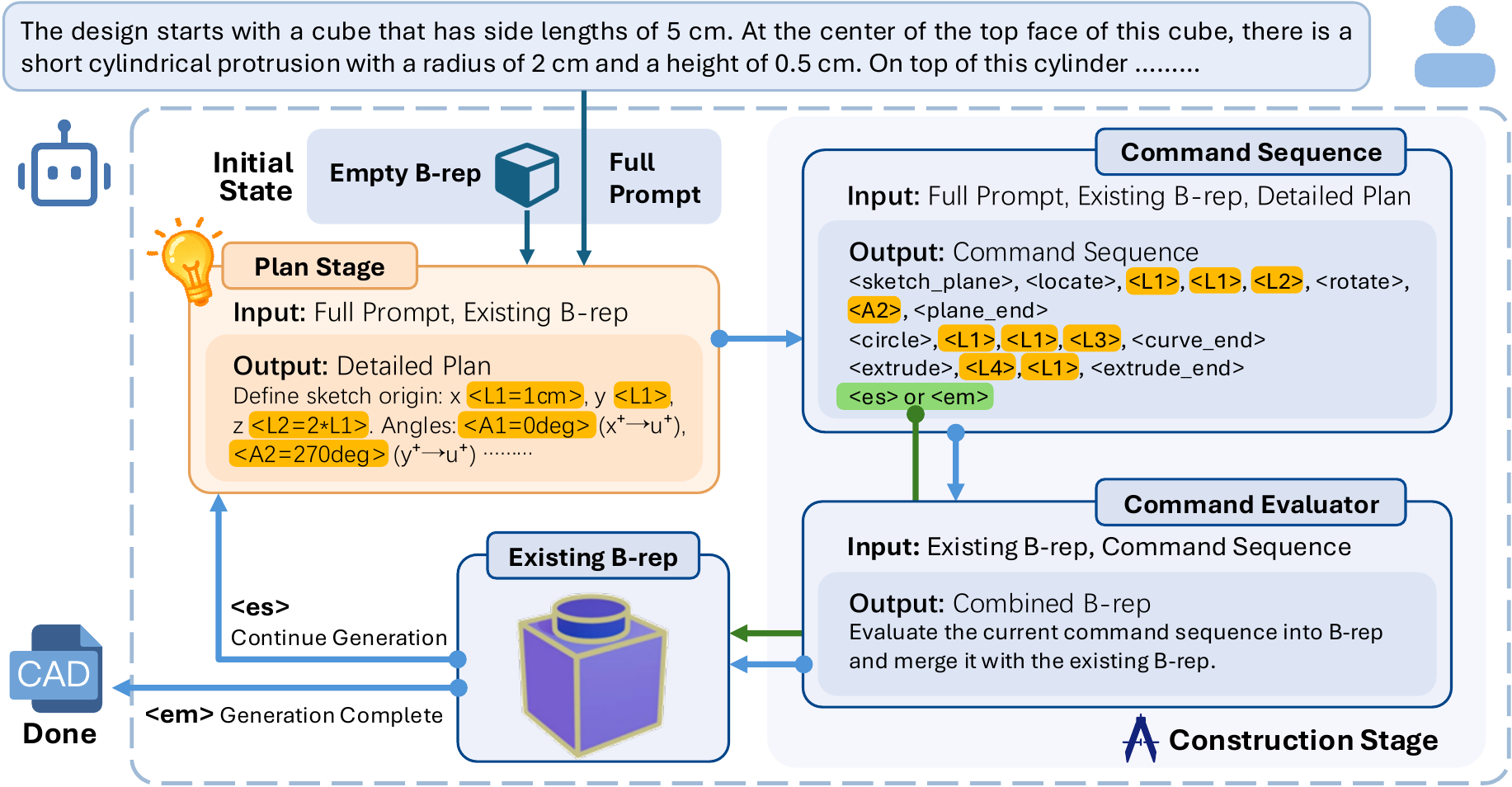}
  \caption{
    \textbf{Pointer-CAD v2 Pipeline.}
    Pointer-CAD v2 generates CAD models in a step-wise manner under a Plan-Then-Construct framework. 
    Each step includes a plan stage and a construction stage.
    The plan stage produces a structured, dimension-aware design plan conditioned on the full prompt and the existing B-rep.
    The construction stage retrieves the planned parameters to form a command sequence, which is constructed to update the B-rep iteratively until completion.
  }
  \label{fig:pipeline}
\end{figure}

To enable precise and flexible CAD generation with explicit dimension-aware parameter reasoning, we propose a \textbf{Plan-Then-Construct} framework built upon Pointer-CAD~\cite{qi2026pointer}.
As illustrated in \cref{fig:pipeline}, the framework decomposes generation into two stages: a plan stage and a construction stage.
The plan stage performs structured reasoning in the textual domain and produces an explicit geometric plan with complete dimensions.
The construction stage converts the plan into executable CAD command sequences.
We first review Pointer-CAD in \cref{sec:preliminary}, and then describe the plan and construction stages in \cref{sec:plan_stage} and \cref{sec:construction_stage}.

\subsection{Preliminary: Pointer-CAD}
\label{sec:preliminary}

Pointer-CAD \cite{qi2026pointer} extends the representation capacity of command sequences by introducing a pointer mechanism that enables direct referencing of geometric entities in the intermediate B-rep.
Instead of generating the entire CAD model in a single forward pass, it adopts a step-wise paradigm to progressively construct the model.
At each step, the model predicts one fundamental operation, including a sketch-extrude pair, a chamfer, or a fillet, conditioned on the current B-rep built from all previous operations.
To support explicit geometric referencing, Pointer-CAD encodes all entities in the current B-rep into a set of embeddings, forming a candidate pool for selection.
During command sequence generation, when an operation needs to reference an existing geometric entity, the language model predicts an embedding vector in the shared space.
The entity with the highest cosine similarity is selected as the output.
This design enables accurate and flexible entity referencing across diverse shapes and topologies, enhancing the representation ability of command sequences for complex CAD modeling.

\subsection{Dimension-Aware Plan Generation}
\label{sec:plan_stage}

\begin{figure}[tb]
  \centering
  \resizebox{\linewidth}{!}{
    \input{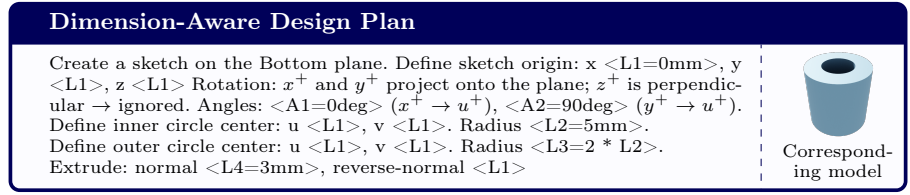}
  }
  \caption{
    \textbf{Example of a dimension-aware design plan.}
    The plan lists geometric parameters in symbolic form.
    Each parameter is enclosed in <> and labeled as \textit{L} for length or \textit{A} for angle.
    All parameters are associated with units and may reference previously defined parameters of the same type.
  }
  \label{fig:plan}
\end{figure}

To enable command sequence representations to express continuous parameters with metric dimensions in the step-wise paradigm, we introduce a structured plan representation in the textual domain.
Unlike conventional command sequence methods~\cite{wu2021deepcad, khan2024text2cad, xu2024cad}, which predict discretized numerical tokens, our plan stage conducts dimension-aware parameter reasoning before command construction. 

At each step, the model generates a structured design plan that specifies all associated geometric parameters in symbolic form. 
As illustrated in \cref{fig:plan}, all referable parameters are enclosed in <> and explicitly typed as either \textit{L} (length) or \textit{A} (angle). 
Each parameter is bound to a unit of measurement (e.g., degrees, meters, millimeters) and supports arithmetic operations as well as references to previously defined parameters of the same type.
This design represents geometric quantities in a metric-scale accurate and non-quantized manner, providing explicit guidance for the subsequent construction stage.

\subsection{Plan-Based Construction Stage}
\label{sec:construction_stage}

During command sequence generation, parameters defined in the plan are retrieved through a similarity-based pointer mechanism.
This improves the dimension precision of the predicted command sequence.

\paragraph{Parameter Extraction and Normalization.}

\begin{figure}[tb]
  \centering
  \includegraphics[width=\linewidth]{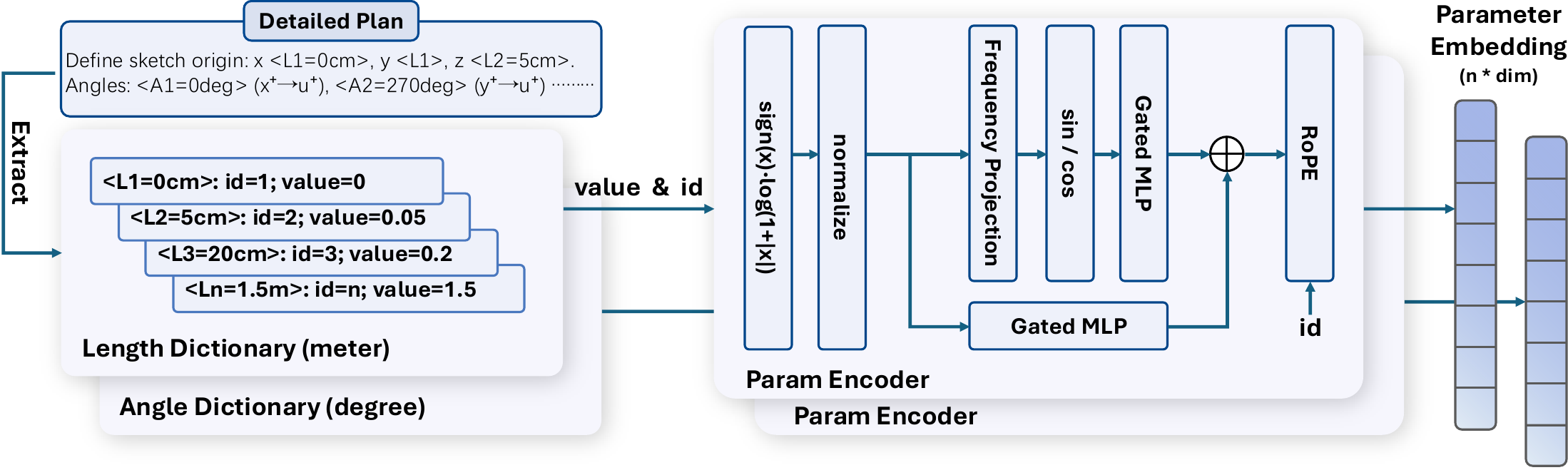}
  \caption{
    \textbf{Plan parameter encoding pipeline.}
    All parameters are first extracted from the generated plan and organized into separate length and angle dictionaries.
    Each dictionary is processed by a parameter encoder that captures both frequency and contextual information.
    The encoder maps every parameter to a dedicated embedding, forming a set of parameter embeddings for subsequent modeling.
  }
  \label{fig:param_encoder}
\end{figure}

As illustrated in \cref{fig:param_encoder}, we first extract all parameters from the generated plan and separate them into length and angle types. 
Each parameter is converted into a unified unit system (meters for length and degrees for angle). 
Since numerical parameter magnitudes may vary significantly after unit conversion, we apply a sign-preserving logarithmic normalization to alleviate scale imbalance:
\begin{equation}
\tilde{x} = \frac{\mathrm{sign}(x) \cdot \log\left(1 + |x|\right)} {\max_{x' \in \mathcal{P}} \left|\mathrm{sign}(x') \cdot \log\left(1 + |x'|\right)\right| + \epsilon} ,
\end{equation}
where $\mathcal{P}$ denotes the parameter set extracted from the plan and $\epsilon$ is a small constant for numerical stability.

\paragraph{Frequency Encoding.}
To enhance representation capacity across diverse numerical ranges, we encode the normalized value $\tilde{x}$ using exponentially increasing frequency bands:
\begin{equation}
\mathbf{z} = \tilde{x} \cdot s \cdot \mathbf{b},
\end{equation}
where $s \in \mathbb{R}$ is a learnable scale and $\mathbf{b} = [2^{0}, 2^{1}, \dots, 2^{K-1}]$ denotes predefined frequency bands.
The resulting Fourier feature~\cite{tancik2020fourfeat, mildenhall2020nerf} is
\begin{equation}
\phi(\tilde{x}) = [\sin(\mathbf{z}), \cos(\mathbf{z})].
\end{equation}

\paragraph{Parameter Embedding.}
To unify dimensionality, both $\tilde{x}$ and $\phi(\tilde{x})$ are projected into a shared embedding space via a gated MLP block $\mathcal{H}(\mathbf{u})$~\cite{dauphin2017language}, defined as:
\begin{equation}
\mathcal{H}(\mathbf{u}) = \mathrm{LayerNorm}\left(\sigma(W_g \mathbf{u}) \odot \mathrm{MLP}(\mathbf{u})\right) ,
\end{equation}
where $\sigma(\cdot)$ denotes the sigmoid function and $\odot$ represents element-wise multiplication. 
The final parameter embedding is computed as
\begin{equation}
\mathbf{e} = \frac{1}{2}\left(\mathcal{H}(\tilde{x}) + \mathcal{H}(\phi(\tilde{x}))\right) .
\end{equation}
To encode parameter identity and positional information, we further apply Rotary Positional Encoding (RoPE)~\cite{su2024roformer} to $\mathbf{e}$.

\paragraph{Similarity-Based Retrieval.}
During command sequence generation, extending Pointer-CAD~\cite{qi2026pointer}, where geometry entity selection activates the pointer to retrieve a target, our mechanism is also triggered whenever a numerical parameter is required, then the LLM predicts an embedding vector $\tilde{\mathbf{e}}$ in the shared embedding space.
We compute cosine similarity between $\tilde{\mathbf{e}}$ and all candidate parameter embeddings ${\mathbf{e}_i}$:
\begin{equation}
\mathrm{similarity}(\mathbf{\tilde{e}}, \mathbf{e}_i) = \frac{\mathbf{\tilde{e}} \cdot \mathbf{e}_i}{|\mathbf{\tilde{e}}| \cdot |\mathbf{e}_i|}.
\end{equation}
The parameter with the highest similarity is selected and inserted into the command sequence.

\paragraph{Progressive Command construction.}
Following~\cite{qi2026pointer}, after similarity-based retrieval, the command sequence is executed immediately to update the current B-rep.
The updated geometry is then fed back to the model to condition the next prediction, enabling progressive construction of the CAD model.
At each step, the plan tokens and the command tokens are generated sequentially within a single forward pass using two modality-specific decoding heads.
A dedicated control token separates the two modalities, ensuring coordinated yet disentangled generation.

Overall, this plan-based progressive command construction strategy ensures that all geometric operations strictly follow the metric scale continuous parameters defined in the plan stage, eliminating quantization errors in conventional command representations.

\section{Experiments}

\subsection{Baselines}
To the best of our knowledge, no prior work explicitly models metric scale geometric parameters for CAD generation.
We therefore adapt two representative methods with different parameter representations as our baselines: the command-sequence-based Pointer-CAD \cite{qi2026pointer} and the code-based CADmium \cite{govindarajan2026cadmium}.

Pointer-CAD generates geometry in a normalized unit cube without explicit units.
For fair comparison under metric scales, we treat it as a $1 \times 1 \times 1$ meter volume and add an autoregressive head to predict a global scaling factor with an $\ell_1$ loss.
In addition, we employ a dedicated text-generation head to predict all involved parameters together with their units in textual form, forming a parameter book.
After rescaling, each geometric parameter is replaced by its nearest entry in the book to maintain consistency between geometry and parameter prediction.
For CADmium, we replace all normalized parameters in the dataset with their original values and units for both training and testing.

In parallel, recent LLMs have shown the ability to perform text-conditioned CAD generation by producing executable CADQuery code.
We additionally evaluate several LLMs under this paradigm, including Qwen3 \cite{yang2025qwen3}, Gemini \cite{google2025gemini3pro}, GPT \cite{openai2025gpt52}, and Claude \cite{anthropic2025claudeopus45}.

\subsection{Dataset}

\begin{figure}[tb]
  \centering
  \includegraphics[width=\linewidth]{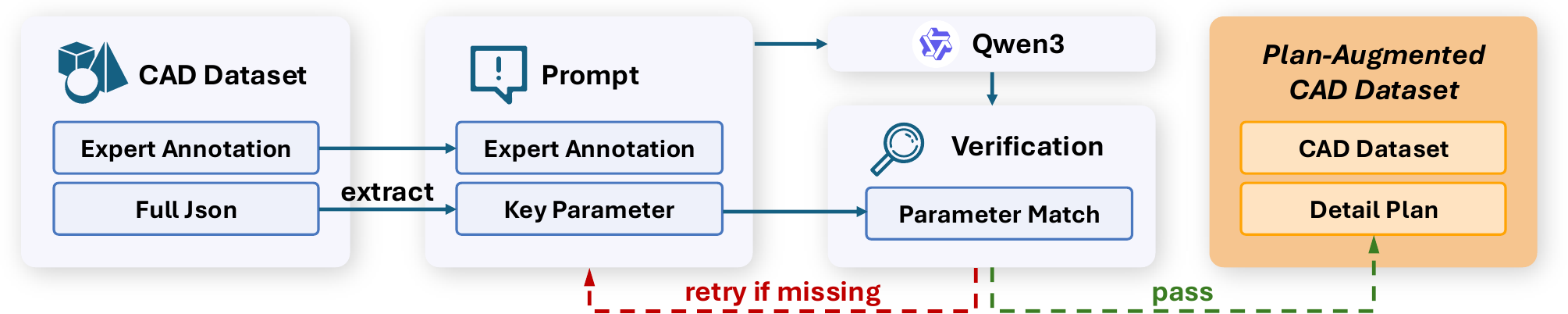}
  \caption{
    \textbf{Plan Construction Pipeline.}
    Key parameters are extracted from raw JSON files and combined with expert annotations to form structured prompts.
    These prompts are fed into Qwen3 to generate detailed design plans.
    The generated plans are verified for parameter completeness.
    If required parameters are missing, Qwen3 is re-prompted up to two times, and any remaining failures are discarded.
  }
  \label{fig:dataset}
\end{figure}

We construct our dataset following the pipeline in \cref{fig:dataset}, based on the datasets released with Pointer-CAD~\cite{qi2026pointer}, which preserve true geometric dimensions without normalization.
Operation-specific parameters are extracted from the JSON files and organized into a compact JSON hint.
The hint, expert annotation, and our designed prompt are fed into Qwen3 to generate a detailed design plan, which is then verified for parameter completeness.
If any parameters are missing, Qwen3 is re-prompted with a non-zero temperature for up to two additional attempts.
The final dataset contains only validated plans.
Further details and data statistics are provided in the supplementary material.

Applying this pipeline to Recap-OmniCAD and its extended version Recap-OmniCAD\textsuperscript{+}, which includes \textit{chamfer} and \textit{fillet} operations, we obtain the plan-augmented datasets OmniCAD-Plan and OmniCAD-Plan\textsuperscript{+}, used for all subsequent training and evaluation.

\subsection{Metrics}

Standard geometric distance metrics, such as Chamfer Distance(CD), are often insufficient for evaluating CAD generation from textual instructions.
CD mainly captures coarse geometric structure, but is less sensitive to fine-grained geometric details (e.g., specific radii or point positions) mentioned in textual prompts and it cannot distinguish whether the underlying B-Rep structure (vertices, edges, and faces) faithfully reconstructs these precise geometric features or merely approximates the overall shape.

To bridge this gap and strictly assess whether the generated models adhere to the dimensional and structural constraints specified in text, we propose three hierarchical metrics: Vertex Accuracy, Edge Accuracy, and Face Accuracy. These metrics shift the evaluation from coarse structural similarity to fine-grained parametric alignment.

For all three metrics, we define the Accuracy Ratio ($Acc$) as:
\begin{equation}
Acc = \frac{N_{match}}{N_{gt}} \times 100
\end{equation}
where $N_{match}$ denotes the number of predicted elements correctly matched with the ground truth (GT), and $N_{gt}$ denotes the total number of corresponding elements in the GT model. Unlike CD, our matching process requires strict parametric consistency:
\begin{itemize}[leftmargin=*, itemsep=2pt]
  \item \textbf{Vertex Accuracy:} This metric evaluates the fidelity of edge endpoints. A predicted vertex is considered correctly matched if its Euclidean distance to the nearest GT point is within a predefined tolerance $\epsilon$. For circular primitives, the center point is utilized as the representative coordinate for matching.
    
  \item \textbf{Edge Accuracy:} An edge is identified as a match only if its endpoints are correctly localized and its underlying \textit{geometric parameters}, (e.g., the center and radius for arcs, or the normal direction for circles) align with the GT within the tolerance $\epsilon$. This metric serves as a direct proxy for the model's capability to adhere to precise dimensional constraints (e.g., ``a cylinder with a 5mm radius'') specified in textual instructions.
    
  \item \textbf{Face Accuracy:} A face is deemed correctly matched if all its constituent boundary edges are successfully identified and its \textit{primitive type} (e.g., plane, cylinder, cone, sphere, or torus) is consistent with the GT. 
\end{itemize}

\begin{table}[t]
  \caption{
    \textbf{Quantitative comparison on the OmniCAD-Plan dataset.}
    Our method consistently outperforms all baselines across three geometric levels.
  }
  \label{tab:comparison-base}
  \centering
  \begin{adjustbox}{max width=0.95\linewidth}
\begin{tabular}{cc @{\hspace{1em}} c @{\hspace{1em}} cccc @{\hspace{1em}} ccc}
\toprule
\multirow{2}[2]{*}{Method} & \multirow{2}[2]{*}{RMR@3} & \multirow{2}[2]{*}{Vertex} & \multicolumn{4}{c}{Edge} & \multicolumn{3}{c}{Face} \\
\cmidrule(r{1em}){4-7} \cmidrule{8-10}  &   &   & Avg & Line  & Circle & Arc & Avg & Plane & Cylinder \\
\midrule
CADmium-0.5B & 78.60 & 76.37 & 74.34 & 76.05 & 72.41 & 4.55 & 70.57 & 71.47 & 66.56 \\
Pointer-CAD-0.5B & 66.03 & 58.87 & 54.81 & 54.30 & 59.18 & 7.34 & 54.70 & 54.44 & 55.88 \\
Ours-0.5B & 87.29 & 90.10 & 86.40 & 85.19 & 92.21 & 64.54 & 84.85 & 83.89 & 89.05 \\
\midrule
CADmium-1.5B & 81.15 & 79.36 & 78.25 & 80.58 & 74.03 & 5.61 & 74.19 & 75.56 & 67.99 \\
Pointer-CAD-1.5B & 72.10 & 64.67 & 62.35 & 62.89 & 62.40 & 24.83 & 61.69 & 62.19 & 59.50 \\
Ours-1.5B & \textbf{91.25} & \textbf{92.76} & \textbf{90.33} & \textbf{89.65} & \textbf{94.23} & \textbf{68.53} & \textbf{89.18} & \textbf{88.73} & \textbf{91.14} \\
\bottomrule
\end{tabular}%

  \end{adjustbox}
\end{table}

\begin{table}[t]
  \caption{
    \textbf{Quantitative comparison on the OmniCAD-Plan\textsuperscript{+} dataset.}
    Our method also outperforms all baselines on the more complex dataset.
  }
  \label{tab:comparison-addon}
  \centering
  \begin{adjustbox}{max width=0.95\linewidth}
\begin{tabular}{cc @{\hspace{1em}} c @{\hspace{1em}} cccc @{\hspace{1em}} ccc}
\toprule
\multirow{2}[2]{*}{Method} & \multirow{2}[2]{*}{RMR@3} & \multirow{2}[2]{*}{Vertex} & \multicolumn{4}{c}{Edge} & \multicolumn{3}{c}{Face} \\
\cmidrule(r{1em}){4-7} \cmidrule{8-10}  &   &   & Avg & Line  & Circle & Arc & Avg & Plane & Cylinder \\
\midrule
CADmium-0.5B & 42.43 & 38.92 & 34.82 & 36.10 & 35.67 & 0.74 & 28.25 & 29.05 & 26.06 \\
Pointer-CAD-0.5B & 54.65 & 47.41 & 42.96 & 44.40 & 43.90 & 5.92 & 42.16 & 43.41 & 38.52 \\
Ours-0.5B & 83.33 & 86.95 & 81.84 & 80.41 & 90.03 & 61.21 & 80.17 & 79.22 & 84.16 \\
\midrule
CADmium-1.5B & 43.41 & 40.15 & 36.96 & 38.88 & 35.88 & 0.28 & 29.80 & 30.99 & 26.15 \\
Pointer-CAD-1.5B & 61.42 & 53.53 & 50.63 & 52.69 & 48.19 & 17.52 & 49.62 & 51.34 & 43.94 \\
Ours-1.5B & \textbf{90.97} & \textbf{92.60} & \textbf{90.13} & \textbf{89.91} & \textbf{94.03} & \textbf{72.29} & \textbf{88.78} & \textbf{88.77} & \textbf{89.64} \\
\bottomrule
\end{tabular}%

  \end{adjustbox}
\end{table}

To maintain consistency across various object scales, the tolerance $\epsilon$ is defined relative to the GT bounding box:
\begin{equation}
\epsilon = 0.001 \times \min \left( \mathbf{b}_{gt}^{max} - \mathbf{b}_{gt}^{min} \right),
\end{equation}
where $\mathbf{b}_{gt}^{min}$ and $\mathbf{b}_{gt}^{max}$ represent the bounding box extrema.
Crucially, all models are evaluated in their original location and scale without normalization, forcing the model to demonstrate a true understanding of metric scale parameters as described in the textual instructions.

We further introduce a repair-aware metric, termed Repairable Model Rate (RMR).
In practical CAD workflows, models with only a few geometric errors can easily be corrected through post-processing.
To reflect this property, a generated model is regarded as repairable under RMR@3 if the number of incorrectly predicted faces does not exceed three.
RMR@3 measures the proportion of such repairable models over the entire test set, and complements strict parametric evaluation by capturing practical usability.

\subsection{Implementation Details}

We use Qwen2.5-0.5B \cite{yang2025qwen3} as the backbone LLM for Pointer-CAD v2 unless stated otherwise.
The model is trained for 10 epochs on NVIDIA H800 GPUs.
Additional implementation details are provided in the supplementary material.

\begin{table}[t]
  \setlength{\tabcolsep}{3pt}
  \caption{
    \textbf{Quantitative comparison under conventional metrics (F1, CD).}
    All methods are evaluated on OmniCAD-Plan with 0.5B models.
    Our method matches Pointer-CAD on Line and Circle F1, which are near saturation, while significantly improving Arc F1.
    CD gains are marginal, as enhanced parameter accuracy is not adequately captured by rough shape-level geometric metrics.
  }
  \label{tab:old_metrics}
  \centering
  \begin{adjustbox}{max width=0.95\linewidth}
\begin{tabular}{ccccccc}
\toprule
Method & Line F1\textuparrow & Arc F1\textuparrow & Circle F1\textuparrow & Extrusion F1\textuparrow & Mean CD\textdownarrow & Median CD\textdownarrow \\
\midrule
CADmium & 85.30 & 14.58 & 76.04 & 88.49 & 7.39 & 0.27 \\
Pointer-CAD & 96.62 & 51.00 & 98.08 & \textbf{99.73} & 3.69 & 0.24 \\
Ours & \textbf{96.75} & \textbf{63.59} & \textbf{98.48} & 99.66 & \textbf{3.15} & \textbf{0.19} \\
\bottomrule
\end{tabular}%

  \end{adjustbox}
\end{table}

\subsection{Comparison with Specialized Text-to-CAD Methods}

In this section, we compare our method with Pointer-CAD~\cite{qi2026pointer} and CADmium~\cite{govindarajan2026cadmium}, under proposed geometry-aware metrics and conventional metrics.

\paragraph{Evaluation on OmniCAD-Plan.}

As shown in \cref{tab:comparison-base}, our method outperforms all baselines on OmniCAD-Plan across all three metric levels. 
The 1.5B model improves average accuracy by 13.49\% over CADmium-1.5B, demonstrating strong parametric precision. 
It also achieves an RMR@3 of 91.25\%, indicating high usability in practical CAD workflows.

\paragraph{Evaluation on OmniCAD-Plan\textsuperscript{+}.}

For OmniCAD-Plan\textsuperscript{+}, we train CADmium on OmniCAD-Plan, as it does not support the additional operations.
Results in \cref{tab:comparison-addon} show that our method consistently outperforms all baselines, demonstrating robustness to increased geometric complexity and richer operations.

\paragraph{Performance under Conventional Metrics.}

Following prior work~\cite{khan2024text2cad, qi2026pointer, govindarajan2026cadmium}, we also report conventional metrics, including F1 score and CD. 
All 0.5B models are evaluated on OmniCAD-Plan. 
As shown in \cref{tab:old_metrics}, while F1 scores for Line and Circle primitives have reached near-saturation, our method still achieves a significant improvement in Arc F1.
Pointer-CAD v2 shows only marginal improvements over Pointer-CAD under CD, which is expected since CD is computed after normalization and is insensitive to small dimensional differences. 
Thus, minor deviations in length or angle have limited impact on shape similarity, and improvements in parameter accuracy are not fully reflected.

\subsection{Comparison with General-Purpose LLMs}

\begin{table}[t]
  \caption{
    \textbf{Quantitative comparison with general-purpose LLMs on OmniCAD-Plan.}
    Our method consistently outperforms all evaluated general-purpose LLMs across the three-level metrics. 
  }
  \label{tab:comparison-LLM-base}
  \centering
  \begin{adjustbox}{max width=0.95\linewidth}
\begin{tabular}{cc @{\hspace{1em}} c @{\hspace{1em}} cccc @{\hspace{1em}} ccc}
\toprule
\multirow{2}[2]{*}{Method} & \multirow{2}[2]{*}{RMR@3} & \multirow{2}[2]{*}{Vertex} & \multicolumn{4}{c}{Edge} & \multicolumn{3}{c}{Face} \\
\cmidrule(r{1em}){4-7} \cmidrule{8-10}  &   &   & Avg & Line  & Circle & Arc & Avg & Plane & Cylinder \\
\midrule
Qwen3-235B-A22B & 58.34 & 53.28 & 49.62 & 48.09 & 56.03 & 30.14 & 46.75 & 46.95 & 45.93 \\
Claude Opus 4.5 & 61.34 & 55.41 & 52.77 & 49.21 & 65.86 & 30.00 & 52.31 & 51.09 & 57.21 \\
GPT-5.2 & 69.16 & 61.03 & 63.55 & 63.11 & 66.59 & 38.36 & 60.52 & 61.41 & 56.94 \\
Gemini 3 Pro & 74.65 & 68.93 & 69.51 & 68.96 & 73.12 & 39.73 & 67.02 & 67.54 & 64.93 \\
Ours-0.5B & 87.77 & 89.24 & 84.34 & 82.17 & 92.61 & 56.16 & 83.19 & 81.88 & 88.44 \\
Ours-1.5B & \textbf{90.89} & \textbf{92.18} & \textbf{89.26} & \textbf{88.21} & \textbf{93.60} & \textbf{67.12} & \textbf{87.93} & \textbf{87.57} & \textbf{89.37} \\
\bottomrule
\end{tabular}%

  \end{adjustbox}
\end{table}

\begin{table}[t]
  \caption{
    \textbf{Quantitative comparison with general-purpose LLMs on OmniCAD-Plan\textsuperscript{+}.}
    On the more complex dataset, our method consistently outperforms all evaluated general-purpose LLMs by a larger margin.
  }
  \label{tab:comparison-LLM-addon}
  \centering
  \begin{adjustbox}{max width=0.95\linewidth}
\begin{tabular}{cc @{\hspace{1em}} c @{\hspace{1em}} cccc @{\hspace{1em}} ccc}
\toprule
\multirow{2}[2]{*}{Method} & \multirow{2}[2]{*}{RMR@3} & \multirow{2}[2]{*}{Vertex} & \multicolumn{4}{c}{Edge} & \multicolumn{3}{c}{Face} \\
\cmidrule(r{1em}){4-7} \cmidrule{8-10}  &   &   & Avg & Line  & Circle & Arc & Avg & Plane & Cylinder \\
\midrule
Qwen3-235B-A22B & 56.18 & 49.01 & 45.37 & 43.07 & 54.21 & 49.62 & 43.51 & 42.57 & 47.36 \\
Claude Opus 4.5 & 61.28 & 54.81 & 52.73 & 48.48 & 67.90 & 66.17 & 53.20 & 51.05 & 62.04 \\
GPT-5.2 & 66.81 & 58.66 & 59.23 & 56.89 & 68.26 & 64.66 & 57.37 & 56.73 & 60.00 \\
Gemini 3 Pro & 74.13 & 67.40 & 68.11 & 66.96 & 73.88 & 55.64 & 66.51 & 66.33 & 67.20 \\
Ours-0.5B & 83.42 & 84.98 & 79.07 & 76.73 & 90.36 & 52.87 & 78.41 & 76.93 & 85.05 \\
Ours-1.5B & \textbf{90.98} & \textbf{91.24} & \textbf{88.02} & \textbf{86.44} & \textbf{95.55} & \textbf{65.41} & \textbf{87.39} & \textbf{86.37} & \textbf{91.73} \\
\bottomrule
\end{tabular}%

  \end{adjustbox}
\end{table}

We randomly sample 2,000 models from the test sets of OmniCAD-Plan and OmniCAD-Plan\textsuperscript{+} and prompt each LLM to generate CadQuery code. 
Results are reported in \cref{tab:comparison-LLM-base} and \cref{tab:comparison-LLM-addon}. 
Among general-purpose models, Gemini-3-Pro performs best on both datasets, yet our method consistently surpasses it. 
Our 1.5B model improves over Gemini-3-Pro by 21.30\% on OmniCAD-Plan and 21.54\% on OmniCAD-Plan\textsuperscript{+} averaged over the three-level metrics. 
General-purpose LLMs also achieve much lower average RMR@3 on both datasets (65.88\% and 64.60\%) than Pointer-CAD v2.
These suggest that current LLMs remain unreliable for CAD generation in terms of parametric precision and robustness.

\begin{table}[t]
  \caption{
    \textbf{Ablation study of the pointer mechanism.}
    The results show that parameter reasoning and plan-based reference are essential, while the pointer mechanism remains robust to scale sensitivity in continuous parameter retrieval.
  }
  \label{tab:ablation-structure}
  \centering
  \begin{adjustbox}{max width=0.95\linewidth}
\begin{tabular}{cc @{\hspace{1em}} c @{\hspace{1em}} cccc @{\hspace{1em}} ccc}
\toprule
\multirow{2}[2]{*}{Variant} & \multirow{2}[2]{*}{RMR@3} & \multirow{2}[2]{*}{Vertex} & \multicolumn{4}{c}{Edge} & \multicolumn{3}{c}{Face} \\
\cmidrule(r{1em}){4-7} \cmidrule{8-10}  &   &   & Avg & Line  & Circle & Arc & Avg & Plane & Cylinder \\
\midrule
w/o Ref. & 65.55 & 66.30 & 58.45 & 55.55 & 71.14 & 17.65 & 56.96 & 54.86 & 66.30 \\
w/o Plan & 80.93 & 76.13 & 73.44 & 73.44 & 75.50 & 29.15 & 72.89 & 73.23 & 71.38 \\
w/o Freq. & 86.77 & 89.02 & 85.20 & 84.10 & 90.67 & 62.24 & 83.68 & 82.85 & 87.31 \\
w/o Norm. & 86.76 & 89.63 & 85.89 & 84.69 & 91.96 & 56.99 & 84.37 & 83.42 & 88.51 \\
Full & \textbf{87.29} & \textbf{90.10} & \textbf{86.40} & \textbf{85.19} & \textbf{92.21} & \textbf{64.54} & \textbf{84.85} & \textbf{83.89} & \textbf{89.05} \\
\bottomrule
\end{tabular}%

  \end{adjustbox}
\end{table}

\subsection{Qualitative Comparison of CAD Generation}

\paragraph{Qualitative Comparison with Baselines.}

\begin{figure}[tb]
  \centering
  \includegraphics[width=\linewidth]{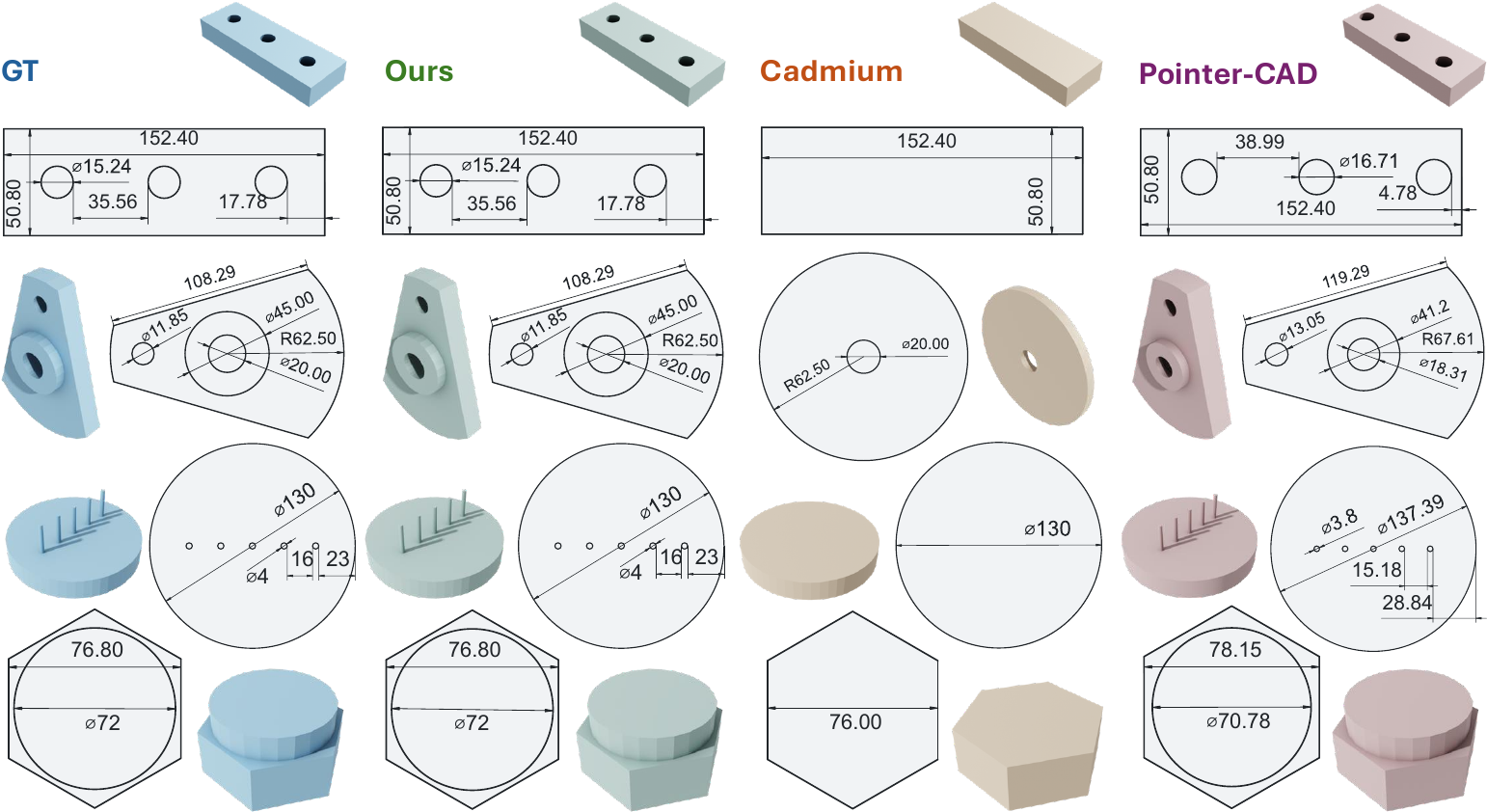}
  \caption{
    \textbf{Qualitative comparison with specialized text-to-CAD methods.}
    Cadmium follows the specified dimensions but shows structural errors.
    Pointer-CAD produces plausible shapes with dimensional deviations.
    Our method achieves both structural correctness and exact dimensional consistency.
  }
  \label{fig:comparison}
\end{figure}

In \cref{fig:comparison}, we present qualitative comparisons of generated CAD models.
The code-based method CADmium \cite{govindarajan2026cadmium} strictly follows numerical dimensions but often produces structural errors, such as missing or incorrectly constructed parts.
Pointer-CAD \cite{qi2026pointer} generates geometrically plausible shapes resembling the ground truth, yet small dimensional deviations remain.
In contrast, our method achieves both structural completeness and exact dimensional consistency. 
The generated models are visually faithful to the target geometry while strictly satisfying all specified parameters.

\paragraph{Complex Generation and Failure Cases.}

\begin{figure}[tb]
  \centering
  \includegraphics[width=0.96\linewidth]{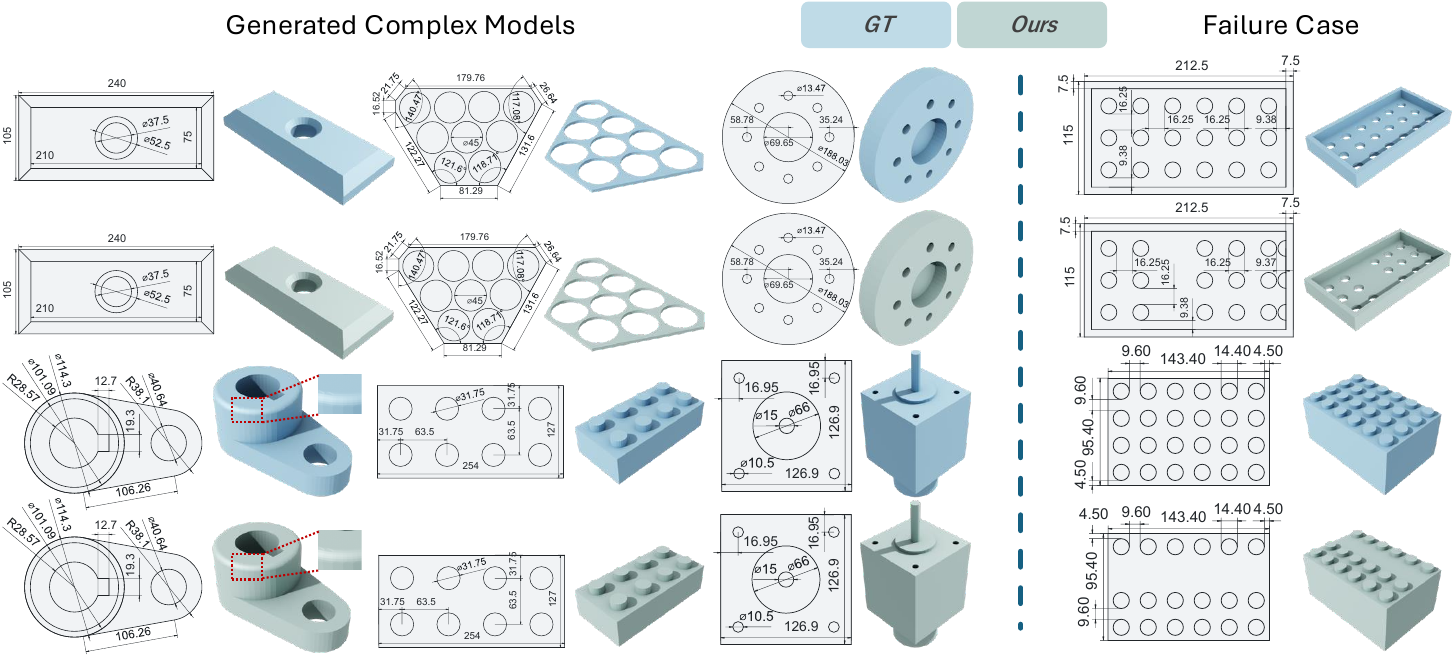}
  \caption{
    \textbf{Visualization of generation and failure cases.}
    Our 1.5B method shows strong robustness across diverse geometric compositions, while occasionally failing to capture extremely fine-grained features.
  }
  \label{fig:visualization}
\end{figure}

\cref{fig:visualization} shows additional CAD models generated by our 1.5B method.
Our approach remains robust across diverse geometries, but occasional inaccuracies appear in very fine features, such as small holes or thin cylinders.
More qualitative results and analysis are provided in the supplementary material.

\subsection{Ablation Study of the Pointer Mechanism}

We ablate the pointer mechanism on OmniCAD-Plan using the 0.5B model.
As shown in \cref{tab:ablation-structure}, we evaluate four variants: w/o Ref., which replaces parameter references with direct regression and nearest-value matching; w/o Plan, which removes the planning stage and directly extracts parameters from the raw prompt; w/o Freq., which removes the frequency encoding mechanism; and w/o Norm., which removes the sign-preserving logarithmic normalization.

The w/o Ref. and w/o Plan variants lead to notable degradations of 26.54\% and 12.96\%, respectively.
This suggests that explicit parameter reasoning and plan-based parameter reference are necessary, as raw prompts often omit computed construction parameters and direct regression is easily affected by nearby values.
In contrast, w/o Freq. and w/o Norm. cause only minor drops of 1.15\% and 0.48\%, respectively, suggesting that the pointer mechanism is robust to scale sensitivity in continuous parameter retrieval.

\subsection{Application: Plan-Based Model Editing}

\begin{figure}[tb]
  \centering
  \includegraphics[width=\linewidth]{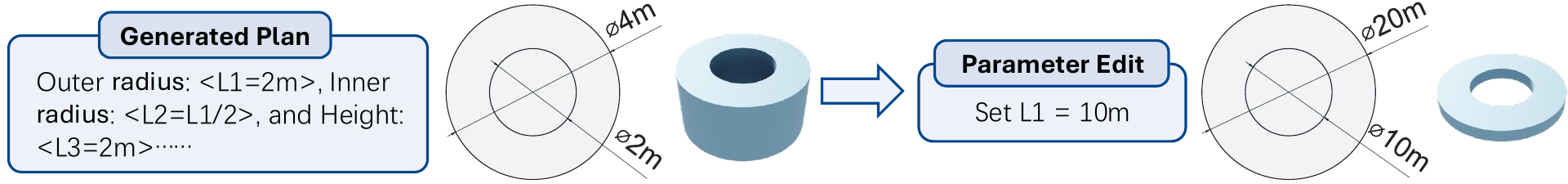}
  \caption{
    \textbf{Example of CAD model editing via plan modification.}
    Changing parameters in the plan adjusts the inner and outer radii while preserving relative proportions.
  }
  \label{fig:plan_editing}
\end{figure}

Our plan representation also enables direct editing of generated CAD models.
Since geometric parameters are explicitly defined in the plan, modifying them updates the model without regenerating the full sequence while preserving relative proportions (\cref{fig:plan_editing}).
Details are provided in the supplementary material.

\section{Conclusion}
In this work, we present Pointer-CAD v2, which addresses quantization limitations of command sequence representations through a \textbf{Plan-Then-Construct} framework.
We generate plan-level annotation, forming a dedicated dataset for structured training.
To better align CAD generation with real-world engineering requirements, we introduce a multi-level accuracy metric that explicitly evaluates geometric and parametric correctness.
Extensive experiments validate the effectiveness of our approach, achieving state-of-the-art performance.

\section{Acknowledgments}

This work is supported in part by the General Research Fund of the Research Grants Council (grant \#17200725), the Shenzhen Loop Area Institute under grants FPF10120250002 and FPF10120260001, and the JC STEM Lab funded by The Hong Kong Jockey Club Charities Trust.

%
%
\bibliographystyle{splncs04}
\bibliography{main}

\clearpage
\vspace*{0em}
{
  \centering \Large \bfseries
  Supplementary Material
  \par
}
\vspace{3em}
\FloatBarrier

\noindent In these supplementary materials, we provide the following:
\begin{itemize}[left=1.5em]
\item Additional experiments and analysis, including further comparisons on the OmniCAD-Plan dataset, comprehensive RMR metrics, extended evaluations under conventional metrics, and an ablation study on the tolerance $\epsilon$ used in our evaluation metric;
\item Additional visualizations, including high-resolution comparisons with baselines, generation results for complex models, representative failure cases, and a video demonstration of the plan-based model editing application;
\item Details of the framework, covering the implementation of the autoregressive decoder, and the training objective;
\item Details of the dataset, including the dataset construction process, dataset statistics, and the annotation prompts.
\end{itemize}
\section{Additional Experiments and Analysis}

\subsection{Additional Results on OmniCAD-Plan}

\begin{table}[t]
  \setlength{\tabcolsep}{3pt}
  \caption{
    \textbf{Quantitative comparison on the OmniCAD-Plan dataset.}
    We evaluate our method against several baselines across model sizes ranging from 0.5B to 7B parameters.
    Our approach consistently outperforms all baselines, with the performance gap widening as model size increases.
  }
  \label{tab:supp-comparison-base}
  \centering
  \begin{adjustbox}{max width=\linewidth}
\begin{tabular}{cc @{\hspace{1em}} c @{\hspace{1em}} cccc @{\hspace{1em}} ccc}
\toprule
\multirow{2}[2]{*}{Method} & \multirow{2}[2]{*}{RMR@3} & \multirow{2}[2]{*}{Vertex} & \multicolumn{4}{c}{Edge} & \multicolumn{3}{c}{Face} \\
\cmidrule(r{1em}){4-7} \cmidrule{8-10}  &   &   & Avg & Line  & Circle & Arc & Avg & Plane & Cylinder \\
\midrule
CADmium-0.5B & 78.60 & 76.94 & 74.44 & 76.16 & 72.45 & 4.55 & 70.62 & 71.52 & 66.56 \\
Pointer-CAD-0.5B & 66.03 & 58.87 & 54.81 & 54.30 & 59.18 & 7.34 & 54.70 & 54.44 & 55.88 \\
Ours-0.5B & 87.29 & 90.10 & 86.40 & 85.19 & 92.21 & 64.54 & 84.85 & 83.89 & 89.05 \\
\midrule
CADmium-1.5B & 81.15 & 79.59 & 78.30 & 80.62 & 74.10 & 5.61 & 74.22 & 75.59 & 68.01 \\
Pointer-CAD-1.5B & 72.10 & 64.67 & 62.35 & 62.89 & 62.40 & 24.83 & 61.69 & 62.19 & 59.50 \\
Ours-1.5B & 91.25 & 92.76 & 90.33 & 89.65 & 94.23 & 68.53 & 89.18 & 88.73 & 91.14 \\
\midrule
CADmium-3B & 82.61 & 81.61 & 80.32 & 82.56 & 76.52 & 6.31 & 76.28 & 77.67 & 70.00 \\
Pointer-CAD-3B & 73.35 & 66.27 & 64.04 & 64.61 & 64.22 & 21.68 & 63.36 & 63.87 & 61.12 \\
Ours-3B & 94.80 & 95.90 & 94.93 & 94.82 & 96.05 & 87.76 & 93.75 & 93.69 & 93.98 \\
\midrule
CADmium-7B & 82.87 & 81.84 & 80.80 & 83.31 & 76.08 & 5.30 & 76.57 & 78.07 & 69.77 \\
Pointer-CAD-7B & 74.33 & 67.59 & 65.43 & 65.98 & 65.37 & 30.07 & 64.63 & 65.14 & 62.37 \\
Ours-7B & \textbf{95.23} & \textbf{96.07} & \textbf{95.13} & \textbf{94.95} & \textbf{96.44} & \textbf{89.86} & \textbf{94.09} & \textbf{93.99} & \textbf{94.53} \\
\bottomrule
\end{tabular}%

  \end{adjustbox}
\end{table}

\cref{tab:supp-comparison-base} presents a comprehensive quantitative comparison on the OmniCAD-Plan dataset between our method and the baselines.
As the model size increases from 0.5B to 7B, the performance gap between our method and CADmium \cite{govindarajan2026cadmium} consistently widens.
Compared to CADmium, the average improvement across the three hierarchical metrics rises from 13.12\% for the 0.5B model to 15.36\% for the 7B model, suggesting that our framework can effectively exploit larger model capacity.
With the 7B model, we achieve an RMR@3 of 95.23\%, indicating that nearly all generated CAD models are usable in practical CAD workflows.

\subsection{Additional Results on RMR Metrics}

\begin{table}[t]
  \setlength{\tabcolsep}{3pt}
  \caption{
    \textbf{Quantitative comparison on the OmniCAD-Plan dataset across RMR metrics (RMR@0 to RMR@5).}
    Our method outperforms all baselines across the full range of RMR metrics, from RMR@0 to RMR@5.
    RMR@3 is the most representative metric for evaluating repairability, as it captures models with only a few sketch or extrusion errors, which can be easily corrected through post-processing.
  }
  \label{tab:supp-rmr}
  \centering
  \begin{adjustbox}{max width=\linewidth}
\begin{tabular}{ccccc @{\hspace{1em}} cccc @{\hspace{1em}} cccc}
\toprule
\multirow{2}[2]{*}{Metric} & \multicolumn{4}{c}{CADmium} & \multicolumn{4}{c}{Pointer-CAD} & \multicolumn{4}{c}{Ours} \\
\cmidrule(r{1em}){2-5} \cmidrule(r{1em}){6-9} \cmidrule{10-13}  & 0.5B & 1.5B & 3B & 7B & 0.5B & 1.5B & 3B & 7B & 0.5B & 1.5B & 3B & 7B \\
\midrule
RMR@0 & 71.63 & 74.24 & 76.17 & 76.16 & 61.26 & 67.66 & 69.10 & 70.14 & 84.36 & 88.45 & 92.42 & \textbf{92.73} \\
RMR@1 & 71.66 & 74.28 & 76.23 & 76.16 & 61.29 & 67.66 & 69.10 & 70.17 & 84.48 & 88.54 & 92.49 & \textbf{92.73} \\
RMR@2 & 72.87 & 75.42 & 77.53 & 77.33 & 62.58 & 68.89 & 70.37 & 71.45 & 85.78 & 89.83 & 93.87 & \textbf{94.21} \\
RMR@3 & 78.60 & 81.15 & 82.61 & 82.87 & 66.03 & 72.10 & 73.35 & 74.33 & 87.29 & 91.25 & 94.80 & \textbf{95.23} \\
RMR@4 & 82.01 & 84.18 & 85.48 & 85.52 & 68.80 & 74.53 & 75.71 & 76.49 & 89.54 & 92.70 & 95.72 & \textbf{96.00} \\
RMR@5 & 85.94 & 87.85 & 88.96 & 88.97 & 78.20 & 81.58 & 82.66 & 83.55 & 93.24 & 95.63 & 97.87 & \textbf{98.09} \\
\bottomrule
\end{tabular}%

  \end{adjustbox}
\end{table}

In the main paper, we evaluate model repairability using RMR@3, which measures the percentage of generated models with at most three incorrectly predicted faces. 
An incorrectly predicted face is defined as a face that misses at least one boundary edge compared to the ground-truth.
From the perspective of sketch and extrusion operations, an incorrect parameter typically affects multiple faces (e.g., side or top faces).
Therefore, RMR@3 approximately corresponds to allowing one minor sketch or extrusion operation error.

In addition to RMR@3, we further evaluate model repairability under a wider range of thresholds, from RMR@0 to RMR@5. 
A model is considered repairable if the number of incorrectly predicted faces does not exceed the given threshold. 
\cref{tab:supp-rmr} reports the performance of different methods under these RMR metrics. 
The RMR@0 to RMR@2 results remain relatively stable.
In contrast, at RMR@5, all models generated by our method achieve scores above 90\%, demonstrating strong robustness for practical CAD workflows.

We focus on RMR@3 in the main paper because it provides the most representative assessment of model repairability. 
RMR@3 reflects models with only a small number of sketch or extrusion errors, which can be easily corrected through post-processing. 
Furthermore, it offers a balanced evaluation of repairability while still distinguishing performance differences between methods.

\subsection{Additional Results on Conventional Metrics}

\begin{table}[t]
  \setlength{\tabcolsep}{3pt}
  \caption{
    \textbf{Quantitative comparison under conventional metrics (F1, CD).}
    All methods are evaluated on the OmniCAD-Plan dataset. Our method outperforms Pointer-CAD and CADmium on most metrics, while CD gains remain marginal, as enhanced parameter accuracy is not adequately captured by rough shape-level geometric metrics.
  }
  \label{tab:supp-old-metrics}
  \centering
  \begin{adjustbox}{max width=\linewidth}
\begin{tabular}{ccccccc}
\toprule
Method & Line F1\textuparrow & Arc F1\textuparrow & Circle F1\textuparrow & Extrusion F1\textuparrow & Mean CD\textdownarrow & Median CD\textdownarrow \\
\midrule
CADmium-1.5B & 84.67 & 24.64 & 76.12 & 88.12 & 8.53 & 0.28 \\
Pointer-CAD-1.5B & \textbf{98.94} & 68.03 & 98.30 & 99.54 & \textbf{2.30} & 0.21 \\
Ours-1.5B & 97.27 & \textbf{94.33} & \textbf{98.81} & \textbf{99.72} & 2.80 & \textbf{0.19} \\
\midrule
CADmium-3B & 90.71 & 38.9 & 82.71 & 92.74 & 5.35 & 0.23 \\
Pointer-CAD-3B & 97.73 & 67.54 & 97.82 & 99.48 & 3.72 & 0.24 \\
Ours-3B & \textbf{99.57} & \textbf{98.55} & \textbf{99.3} & \textbf{99.83} & \textbf{1.13} & \textbf{0.17} \\
\midrule
CADmium-7B & 93.28 & 40.51 & 84.44 & 95.38 & 4.68 & 0.23 \\
Pointer-CAD-7B & 98.66 & \textbf{94.89} & 98.08 & 99.43 & 2.14 & 0.2 \\
Ours-7B & \textbf{99.23} & 94.8 & \textbf{99.13} & \textbf{99.75} & \textbf{1.02} & \textbf{0.17} \\
\bottomrule
\end{tabular}%

  \end{adjustbox}
\end{table}

\begin{table}[t]
  \setlength{\tabcolsep}{3pt}
  \caption{
    \textbf{CD results on a 2K subset of the OmniCAD-Plan dataset.}
    Pointer-CAD v2 achieves lower CD than general-purpose LLMs.
  }
  \label{tab:supp-old-metrics-llm}
  \centering
  \begin{adjustbox}{max width=\linewidth}
\begin{tabular}{ccccccc}
\toprule
  & GPT-5.2 & Gemini 3 Pro & Claude Opus 4.5 & Qwen3 & Ours-0.5B & Ours-1.5B \\
\midrule
Mean CD\textdownarrow & 12.01 & 12.78 & 15.21 & 19.77 & 1.99 & \textbf{0.91} \\
Median CD\textdownarrow & 0.24 & 0.24 & 0.26 & 0.34 & \textbf{0.19} & \textbf{0.19} \\
\bottomrule
\end{tabular}%

  \end{adjustbox}
\end{table}

\cref{tab:supp-old-metrics} compares our method with Pointer-CAD and CADmium under conventional metrics.
Our method outperforms both baselines on most metrics.
In the few cases where Pointer-CAD performs better, the gains are marginal and the corresponding metrics are close to saturation, suggesting that these coarse geometric metrics do not fully reflect improvements in parameter precision.

\cref{tab:supp-old-metrics-llm} reports CD results on a 2K subset of OmniCAD-Plan, comparing Pointer-CAD v2 with general-purpose LLMs.
Pointer-CAD v2 achieves the lowest CD and substantially outperforms general-purpose LLMs in Mean CD, indicating better geometric accuracy and stronger robustness on challenging cases.

Overall, conventional metrics provide useful evidence of geometric fidelity and show a trend consistent with our proposed metrics.
However, they remain insufficient to capture improvements in parameter accuracy, highlighting the need for specialized metrics such as RMR for evaluating CAD model generation.

\subsection{Ablation Study on Metric Tolerance}

\begin{figure}[t]
  \centering
  \includegraphics[width=0.8\linewidth]{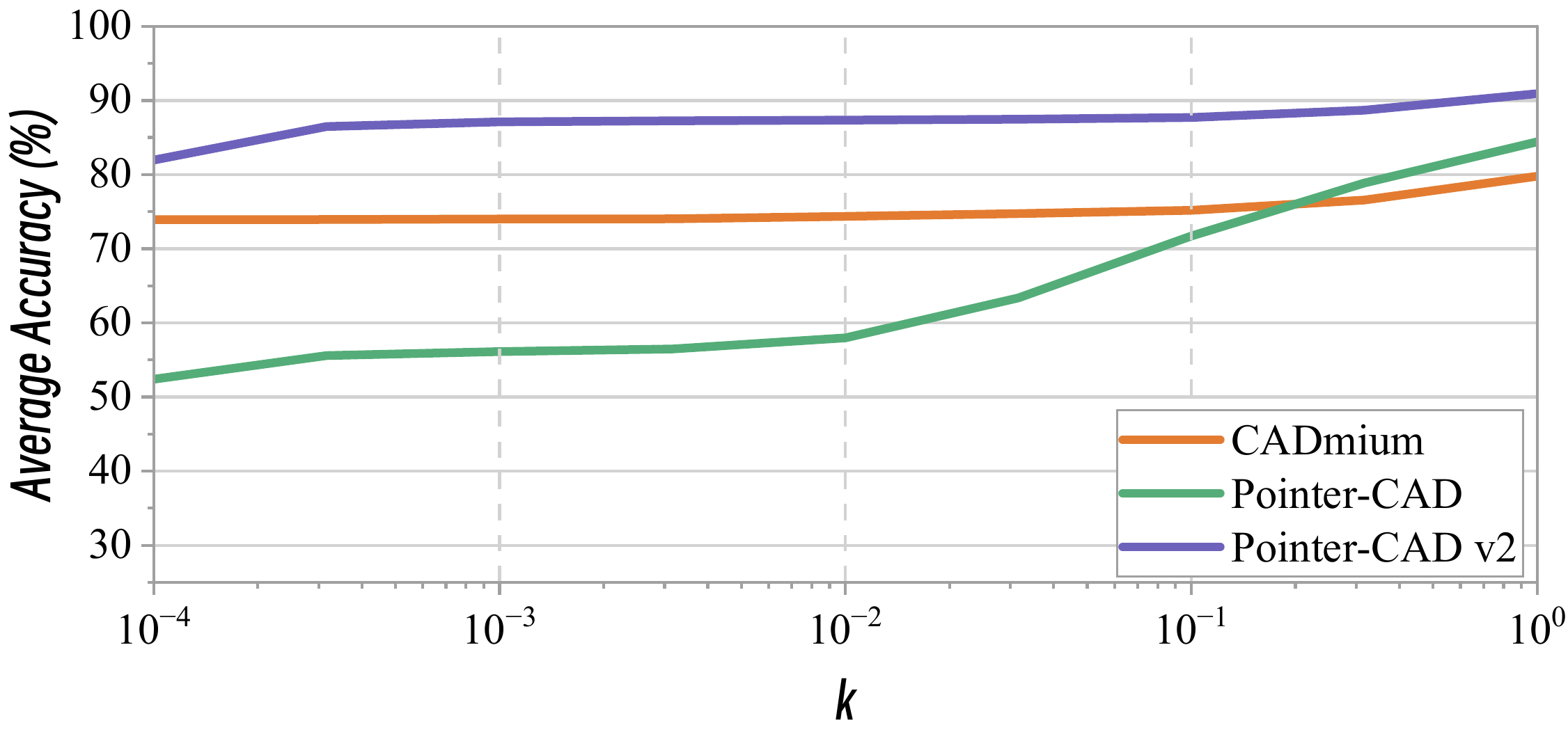}
  \caption{
    \textbf{Ablation study on metric tolerance.}
    We assess the average accuracy across three metric levels for different values of the tolerance factor $k$.
    Since performance remains stable when $k$ is around $10^{-3}$, we set $k=10^{-3}$ in all experiments.
  }
  \label{fig:supp_k_ablation}
\end{figure}

As defined in Eq.~8 of the main paper, the tolerance $\epsilon$ is
\begin{equation}
\epsilon = k \times \min \left( \mathbf{b}_{gt}^{max} - \mathbf{b}_{gt}^{min} \right),
\end{equation}
where $k=0.001$.
To better understand the impact of $k$ on the evaluation results, we conduct an ablation study by varying this factor.

As shown in \cref{fig:supp_k_ablation}, we evaluate the average accuracy across the three metric levels for different values of $k$.
When $k$ is above $10^{-2}$, the performance of all methods increases substantially, especially for Pointer-CAD, since a larger tolerance makes the evaluation more lenient and allows greater deviations from the ground-truth geometry.
In contrast, when $k$ decreases to $10^{-4}$, both Pointer-CAD and Pointer-CAD v2 show a performance drop.
This occurs because, during dataset construction, the plans in the Pointer-CAD v2 dataset and the parameter books in the Pointer-CAD dataset both involve multiple format conversions of the parameters.
These conversions may introduce small numerical errors, but their impact is negligible in practice.
For CADmium \cite{govindarajan2026cadmium}, which is a code-based method, no such conversions are required, and its performance therefore remains relatively stable.

To balance strictness and robustness, we set $k=10^{-3}$ in the main paper.
This value provides a reasonable tolerance for evaluating metric-scale accuracy while still distinguishing different levels of geometric fidelity.

\section{Additional Visualizations}

\subsection{Qualitative Comparisons and Failure Cases}

We provide an interactive \textcolor{red}{HTML visualization (\textbf{visualization.html})} in the supplementary zip file, which includes qualitative comparisons with the baselines, generations of complex models, and representative failure cases.

These failures reflect error propagation in the Plan-Then-Construct paradigm, where errors in planning, referencing, or construction cause missing or misplaced features in the final output.
Most fine-grained features in main-paper Fig. 7 are correctly constructed, indicating that specified values are generally retrieved and used, while the few failures stem from propagated planning or parameter-extraction errors.

\subsection{Additional Details on Plan-Based Model Editing}

After generating the plan and the command sequence, the parameter books are extracted from the plan and automatically evaluated to replace the corresponding parameters in the command sequence. 
By modifying the parameters specified in the plan, we can directly edit the parameter books while preserving the indices and semantic meanings of the parameters. 
The edited CAD model is then obtained by evaluating the command sequence with the modified parameter books. 
\textcolor{red}{A demonstration video (\textbf{plan\_based\_model\_editing.mp4})} is also provided in the supplementary zip file, illustrating the plan-based model editing process.
\section{Details of the framework}

\subsection{Implementation Details of the Autoregressive Decoder}

\begin{table}[t]
  \setlength{\tabcolsep}{3pt}
  \caption{
    \textbf{Label Token Definitions.}
    This table provides a comprehensive list of all \textit{label tokens} used in our command sequence representation.
  }
  \label{tab:supp-notation}
  \centering
  \begin{adjustbox}{max width=\linewidth}
\begin{tabular}{cccc}
\toprule
Notation & ID & Related Tokens & Description \\
\midrule
\commandnotation{em} & 1 & - & End of model (this step is the final step of the model) \\
\commandnotation{es} & 2 & - & End of step (additional steps are required after this one) \\
\commandnotation{ss} & 3 & - & Start of sketch \\
\commandnotation{se} & 4 & - & Start of extrusion \\
\commandnotation{sc} & 5 & - & Start of chamfer \\
\commandnotation{sf} & 6 & - & Start of fillet \\
\commandnotation{sp} & 7 & - & Start of profile \\
\commandnotation{sl} & 8 & - & Start of loop \\
\commandnotation{sx} & 9 & - & Start of curve \\
\commandnotation{pe} & 10 & Pointer Token & Pointer to an Edge or Face \\
\commandnotation{pd} & 11 & Pointer Token & Empty pointer \\
\commandnotation{or} & \{12, 13\} & - & Orientation (Clockwise, Counter-Clockwise) \\
\commandnotation{dr} & $[14,\, 19]$ & - & Direction (X+, X-, Y+, Y-, Z+, Z-) \\
\commandnotation{bo} & $[20,\, 23]$ & - & Boolean (New, Join, Cut, Intersect) \\
\textbf{\commandnotation{lv}} & \textbf{24} & \textbf{Value Token} & \textbf{Pointer to a Length} \\
\textbf{\commandnotation{av}} & \textbf{25} & \textbf{Value Token} & \textbf{Pointer to an Angle} \\
\bottomrule
\end{tabular}%

  \end{adjustbox}
\end{table}

We follow the Pointer-CAD framework \cite{qi2026pointer} to convert LLM outputs into the command sequence.
At each autoregressive decoding step, the last hidden state of the transformer decoder is used to predict the next token in the command sequence.
In Pointer-CAD, a single head jointly decodes label and value tokens.
However, in Pointer-CAD v2, we separate this process by introducing three distinct heads: \textbf{a Label Head, a Value Head, and a Pointer Head}, each responsible for predicting a specific type of token.

The \textbf{Label Head} is a classification head that predicts \textit{label tokens}.
Its output is one of the \textit{label tokens} defined in Table~\ref{tab:supp-notation}, which can have the following meanings:
\begin{itemize}
\item A token without related tokens, which provides semantic information for the command sequence.
\item \commandnotation{lv} and \commandnotation{av}, which are related with value tokens. 
When the Label Head predicts these tokens, it indicates that the current token corresponds to a length or angle value. 
In this case, the output from the \textbf{Value Head} is used, and the candidate dictionary is selected according to \commandnotation{lv} (length value dictionary) or \commandnotation{av} (angle value dictionary).
\item \commandnotation{pe} and \commandnotation{pd}, which are related with pointer tokens. 
When the model predicts \commandnotation{pe}, it indicates that the current token is a pointer, and the output from the \textbf{Pointer Head} is used.
\end{itemize}

Both the \textbf{Value Head} and \textbf{Pointer Head} are linear layers that output a 128-dimensional vector.
This vector is then used to perform a similarity search (via cosine similarity) against the embeddings of the candidate representations with the same dimensionality.
The \textbf{Value Head} is responsible for the embeddings of either length or angle values, selected by \commandnotation{lv} and \commandnotation{av}, respectively.
The \textbf{Pointer Head} handles the embeddings of geometric entities (faces and edges) from the existing B-rep.

\subsection{Details of Training Objective}

Following Pointer-CAD~\cite{qi2026pointer}, the overall training objective $\mathcal{L}$ combines the cross-entropy losses for plan tokens ($\mathcal{L}_t$) and label tokens ($\mathcal{L}_l$), together with the contrastive losses for value tokens ($\mathcal{L}_v$) and pointer tokens ($\mathcal{L}_p$):
\begin{equation}
\mathcal{L} = \lambda_t \cdot \mathcal{L}_t + \lambda_l \cdot \mathcal{L}_l + \lambda_v \cdot \mathcal{L}_v + \lambda_p \cdot \mathcal{L}_p .
\end{equation}
The hyperparameters $\lambda_t$, $\lambda_l$, $\lambda_v$, and $\lambda_p$ control the relative contribution of each loss term.
In all experiments, we set $\lambda_t = 0.2$, $\lambda_l = 0.3$, $\lambda_v = 0.2$, and $\lambda_p = 0.3$.
Higher weights are assigned to the label and pointer losses because these tokens play a critical role in accurate command generation.
\section{Details of the Dataset}

\subsection{Dataset Construction}

The detailed design plans in the OmniCAD-Plan and OmniCAD-Plan\textsuperscript{+} datasets are automatically constructed using Qwen3-32B \cite{yang2025qwen3}. 
Specifically, we design a detailed prompt that describes the CAD model and asks the Qwen3 to generate the corresponding design plan. 
The generated plans are automatically corrected for potential syntax errors and then verified through programmatic validation and manual inspection. 
This process ensures that the generated plans faithfully reflect the actual modeling procedures of the corresponding CAD models.
In this section, we introduce the plan correction and validation processes.

\paragraph{Plan Correction}

The plan correction process identifies and resolves syntax errors in the generated plans, including missing or mismatched parameter IDs, duplicated parameter IDs or values, and incorrect parameter types. 
We correct these issues according to the syntax rules of the modeling operations to ensure that parameter IDs are consecutive and parameter values remain unique within each plan.
This procedure is fully automated and applied to all generated plans, producing syntactically valid plans for subsequent validation.

\paragraph{Plan Validation}

\begin{figure}[t]
  \centering
  \includegraphics[width=0.8\linewidth]{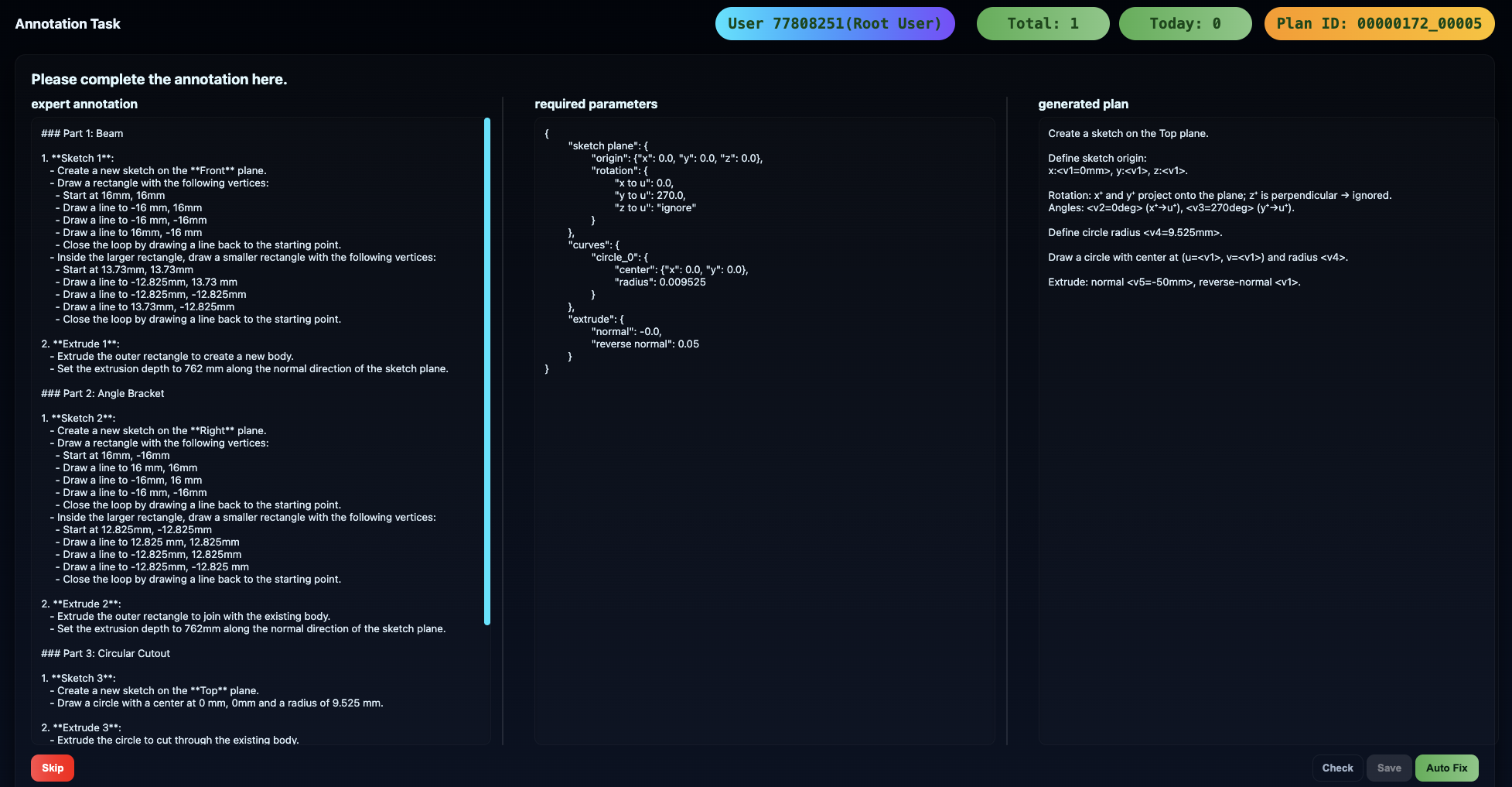}
  \caption{
    \textbf{GUI platform for manual inspection.}
    This interface is used for manual inspection of the generated plans.
    Annotators can review expert annotations, required parameters with their values, and the generated plans.
    They can also directly revise the plans to ensure accuracy and consistency.
  }
  \label{fig:supp_web}
\end{figure}

The plan validation process consists of two stages: programmatic validation and manual inspection. 
During programmatic validation, we extract the required parameters from the source dataset and verify whether the generated plans include all necessary parameters. 
This check is performed automatically for all generated plans.
Plans that fail this stage are sent back to Qwen3 for regeneration, while plans that pass proceed to manual inspection.
To ensure annotation quality, we further perform manual inspection of the generated plans.
Before the inspection, annotators are required to complete a set of sample questions to familiarize themselves with the guidelines.
During inspection, they review the expert annotations, required parameters with their values, and the generated plans in a GUI platform (see \cref{fig:supp_web}).
Annotators can directly revise the plans when necessary, which helps ensure accuracy and consistency.

\subsection{Dataset Statistics}

\begin{figure}[t]
  \centering
  \begin{subfigure}{0.48\linewidth}
    \includegraphics[width=\linewidth]{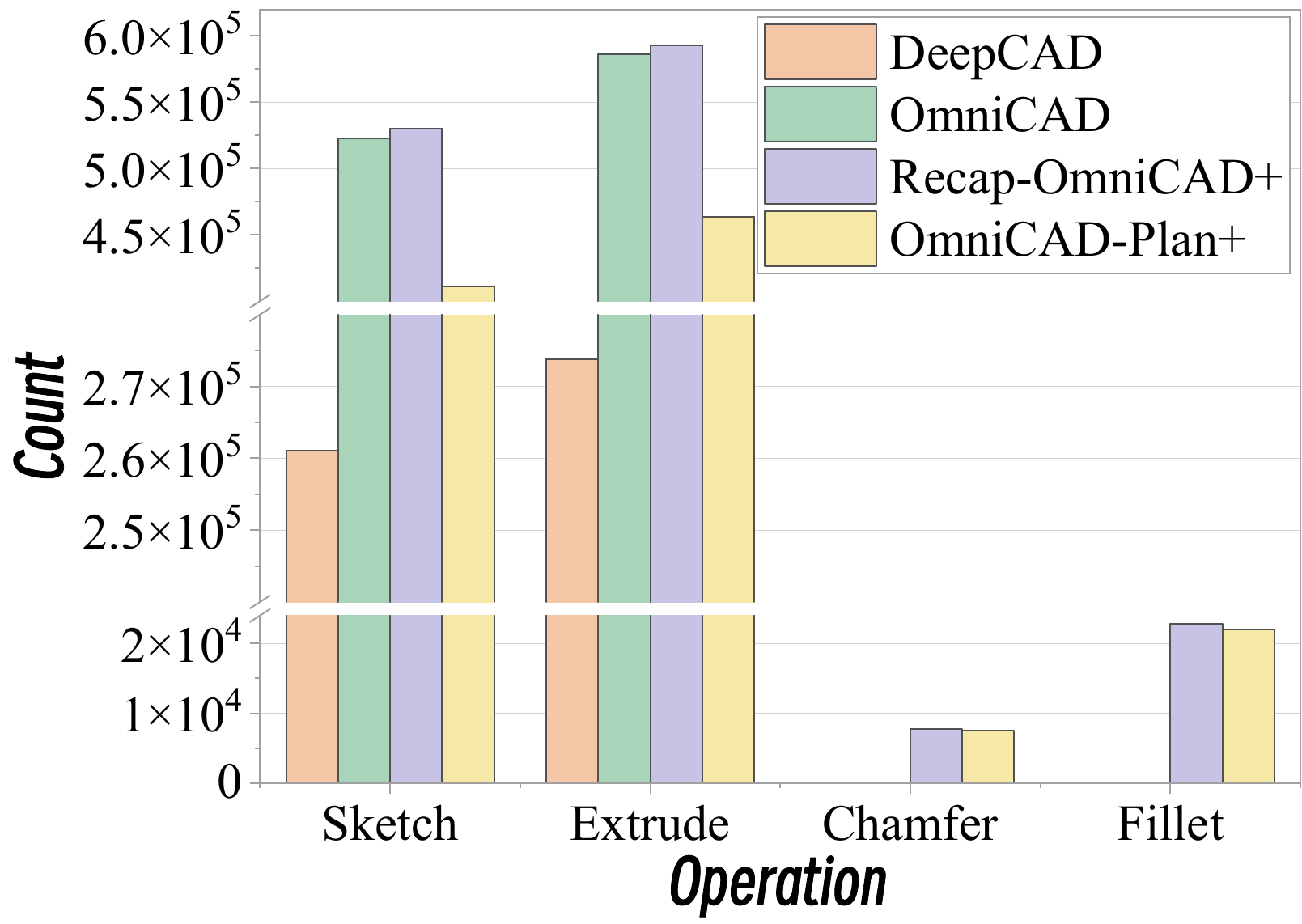}
    \caption{
      \textbf{Distribution of modeling operations across datasets.}
      This figure shows the number of modeling operations (e.g., sketch, extrude, chamfer, and fillet) in different datasets.
    }
    \label{fig:supp_data_feature}
  \end{subfigure}
  \hfill
  \begin{subfigure}{0.48\linewidth}
    \includegraphics[width=\linewidth]{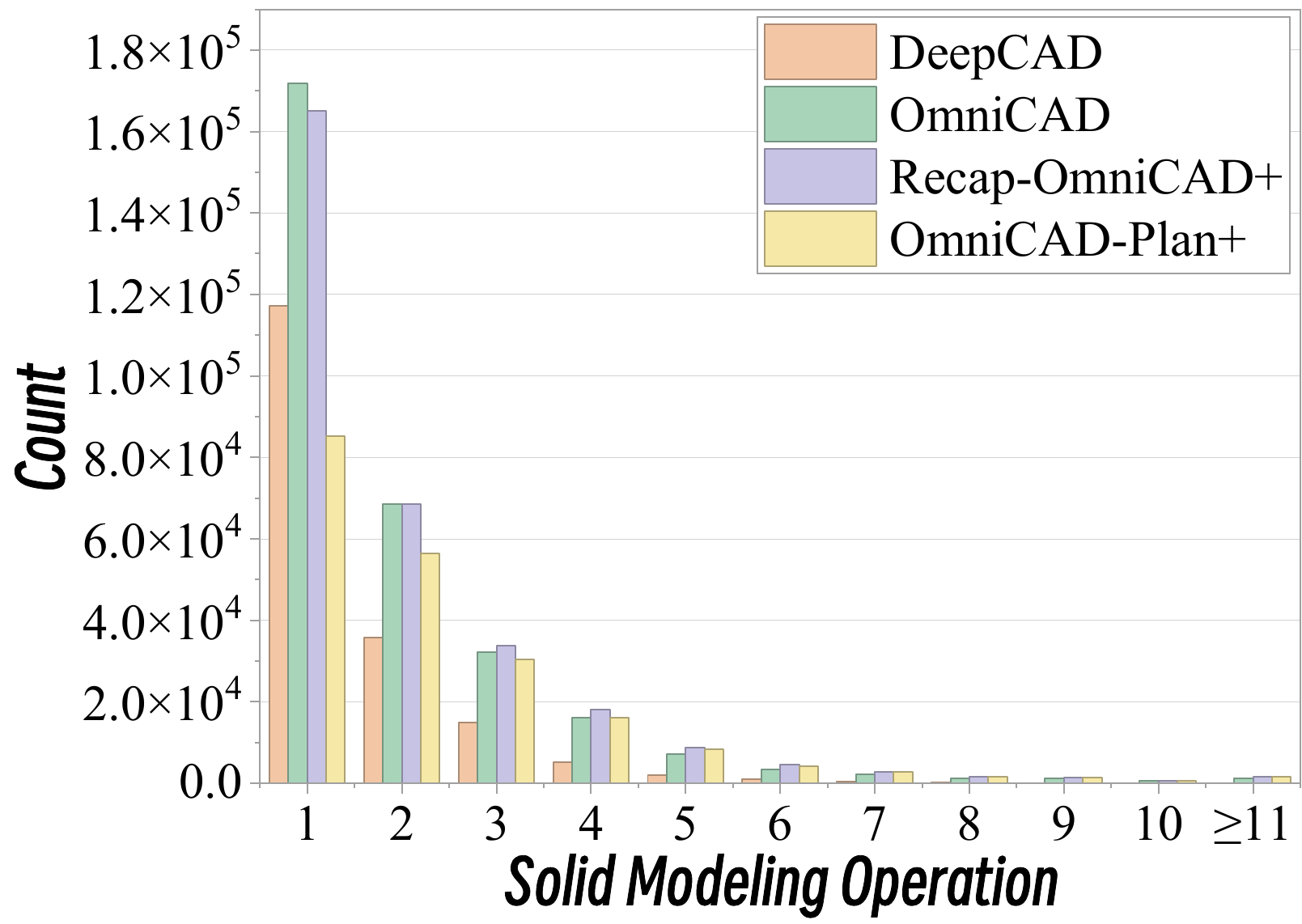}
    \caption{
      \textbf{Distribution of modeling steps per CAD model across datasets.}
      This figure compares the number of solid modeling operations required per model in each dataset.
    }
    \label{fig:supp_data_step}
  \end{subfigure}
  \caption{
    \textbf{Dataset statistics.}
    We analyze the distribution of modeling operations and the number of modeling steps per CAD model across multiple datasets, including DeepCAD \cite{wu2021deepcad}, OmniCAD \cite{xu2024cad}, Recap-OmniCAD\textsuperscript{+} \cite{qi2026pointer}, and OmniCAD-Plan\textsuperscript{+}.
  }
  \label{fig:supp_data}
\end{figure}

\begin{figure}[t]
  \centering
  \begin{subfigure}{0.48\linewidth}
    \centering
    \includegraphics[width=0.9\linewidth]{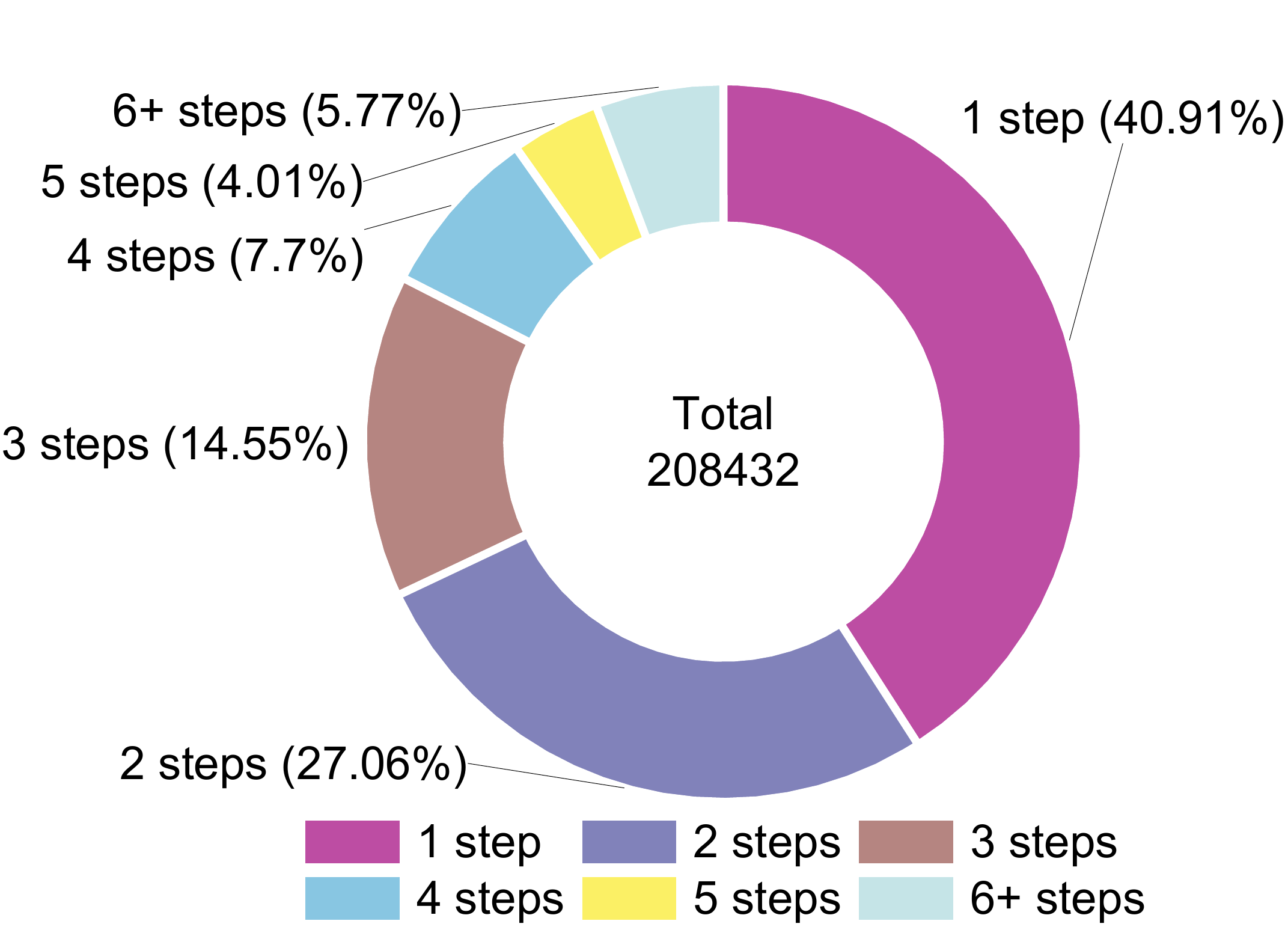}
    \caption{
      \textbf{Distribution of modeling steps in OmniCAD-Plan\textsuperscript{+}.}
      40.91\% of the models require only one modeling step.
    }
    \label{fig:supp_long_tail_ours}
  \end{subfigure}
  \hfill
  \begin{subfigure}{0.48\linewidth}
    \centering
    \includegraphics[width=0.9\linewidth]{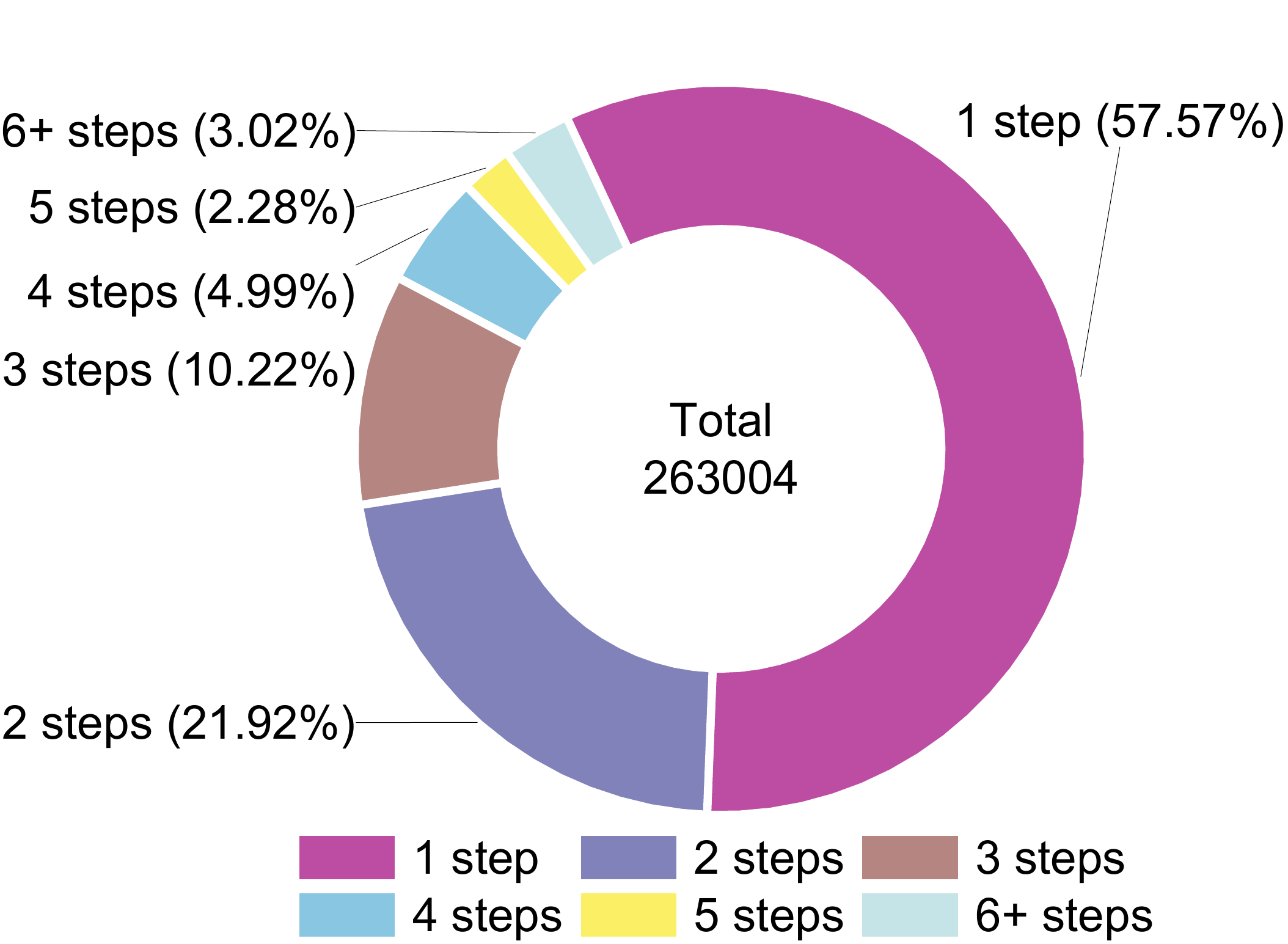}
    \caption{
      \textbf{Distribution of modeling steps in other CAD datasets.}
      On average, 57.57\% of the models require only one modeling step.
    }
    \label{fig:supp_long_tail_others}
  \end{subfigure}
  \caption{
    \textbf{Distribution of modeling steps.}
    All datasets show a clear long-tail distribution, with many models requiring only one modeling step.
  }
  \label{fig:supp_long_tail}
\end{figure}

We conduct a comprehensive statistical analysis of the distribution of modeling operations and the number of modeling steps per CAD model across multiple datasets, as shown in \cref{fig:supp_data}.

\cref{fig:supp_data_feature} presents the distribution of modeling operations across datasets.
Compared with Recap-OmniCAD\textsuperscript{+} \cite{qi2026pointer}, OmniCAD-Plan\textsuperscript{+} exhibits a substantial reduction in sketch and extrude operations, while chamfer and fillet operations remain largely stable.
On average, sketch and extrude operations decrease by 22.08\% relative to Recap-OmniCAD\textsuperscript{+}, whereas chamfer and fillet operations decrease by only 3.7\%.
We attribute this difference to the higher complexity of sketch-extrude pairs.
Such operations are usually associated with complex sketches containing multiple curves, which require a larger number of locate steps and therefore increase the difficulty of generating valid plans with Qwen3.

\cref{fig:supp_data_step} illustrates the distribution of modeling steps per CAD model across different datasets.
A clear long-tail distribution is observed in all datasets, where a substantial proportion of models require only a single step.
In OmniCAD-Plan\textsuperscript{+}, one-step models account for 40.91\% of the dataset (see \cref{fig:supp_long_tail_ours}).
By comparison, other datasets exhibit a higher proportion of one-step models, with an average of 57.57\% (see \cref{fig:supp_long_tail_others}).
Interestingly, although some models are removed due to invalid plans, the long-tail effect in the OmniCAD-Plan\textsuperscript{+} series becomes less pronounced instead of stronger.

\subsection{Details of Annotation Prompts}

During dataset construction, we design different annotation prompts for different operation types to guide Qwen3 \cite{yang2025qwen3} in generating accurate and relevant design plans.

For the system prompt, we provide a detailed description of each modeling operation. 
Examples are shown in \cref{fig:supp_data_sketch_extrude} for the sketch-extrude operation pair, \cref{fig:supp_data_chamfer} for the chamfer operation, and \cref{fig:supp_data_fillet} for the fillet operation. 
These prompts explain the functionality of each operation and clarify the expected structure of the generated plan.

For the user prompt, we adopt a unified template that contains the parameters required by each operation type together with their corresponding values extracted from the source dataset. 
As illustrated in \cref{fig:supp_data_user}, the template provides the necessary information for Qwen3 to understand the CAD model. 
The placeholder \texttt{INSERT\_PROMPT\_HERE} is replaced with the expert-level model description from the source dataset. 
The placeholder \texttt{INSERT\_JSON\_HERE} is replaced with the required parameters and their corresponding values for the specific operation type in JSON format.

Examples of the JSON parameters for different operation types are shown in 
\cref{fig:supp_data_json_sketch_extrude,fig:supp_data_json_chamfer_fillet}. 
Each JSON entry specifies the required parameters and their corresponding values extracted from the source dataset. 
These structured parameters provide precise geometric and dimensional information, enabling Qwen3 to generate plans that are consistent with the original CAD model.

\begin{figure}[tb]
  \centering
  \resizebox{\linewidth}{!}{
    \input{supp/figures/prompts/sketch_extrude}
  }
  \caption{
    \textbf{System prompt for sketch-extrude operation pairs.}
    This prompt provides detailed descriptions of sketch and extrude operations.
  }
  \label{fig:supp_data_sketch_extrude}
\end{figure}

\begin{figure}[tb]
  \centering
  \resizebox{\linewidth}{!}{
    \input{supp/figures/prompts/chamfer}
  }
  \caption{
    \textbf{System prompt for chamfer operation.}
    This prompt provides a detailed description of the chamfer operation and specifies the expected structure of the generated design plan.
  }
  \label{fig:supp_data_chamfer}
\end{figure}

\begin{figure}[tb]
  \centering
  \resizebox{\linewidth}{!}{
    \input{supp/figures/prompts/fillet}
  }
  \caption{
    \textbf{System prompt for fillet operation.}
    This prompt provides a detailed description of the fillet operation and specifies the expected structure of the generated design plan.
  }
  \label{fig:supp_data_fillet}
\end{figure}

\begin{figure}[tb]
  \centering
  \resizebox{\linewidth}{!}{
    \input{supp/figures/prompts/user}
  }
  \caption{
    \textbf{User prompt.}
    This prompt specifies the parameters that should appear in the generated design plan.
  }
  \label{fig:supp_data_user}
\end{figure}

\begin{figure}[tb]
  \centering
  \resizebox{\linewidth}{!}{
    \input{supp/figures/jsons/sketch_extrude}
  }
  \caption{
    \textbf{JSON parameters for sketch-extrude operation pairs.}
    This figure shows the required parameters and their corresponding values for sketch and extrude operations in JSON format.
  }
  \label{fig:supp_data_json_sketch_extrude}
\end{figure}

\begin{figure}[tb]
  \centering
  \resizebox{\linewidth}{!}{
    \input{supp/figures/jsons/chamfer_fillet}
  }
  \caption{
    \textbf{JSON parameters for chamfer and fillet operations.}
    This figure shows the required parameters and their corresponding values for chamfer and fillet operations in JSON format.
  }
  \label{fig:supp_data_json_chamfer_fillet}
\end{figure}

\end{document}